\documentclass[acmtog]{acmart}

\AtBeginDocument{%
  }

\setcopyright{acmlicensed}
\copyrightyear{2025}
\acmYear{2025}
\acmDOI{XXXXXXX.XXXXXXX}

\acmConference[Conference acronym 'XX]{Make sure to enter the correct
  conference title from your rights confirmation emai}{June 03--05,
  2018}{Woodstock, NY}
\acmISBN{978-1-4503-XXXX-X/18/06}

\newif\ifSkeleton
\Skeletontrue 
\newcommand{\ModiColor}{black} 

\newcommand{\Gond}[1]{\ifSkeleton \textcolor{\ModiColor}{#1} \fi}
\newcommand{\Manu}[1]{\ifSkeleton #1 \fi}

\begin{document}

\title{Real-Time Position-Aware View Synthesis from Single-View Input}

\author{Manu Gond}
\email{manu.gond@miun.se}
\orcid{0009-0006-9845-1652}
\affiliation{%
  \institution{Mid Sweden University}
  \city{Sundsvall}
  \country{Sweden}
}
\affiliation{%
  \institution{Technical University of Berlin}
  \city{Berlin}
  \country{Germany}
}

\author{Emin Zerman}
\orcid{0000-0002-3210-8978}
\email{emin.zerman@miun.se}
\affiliation{%
  \institution{Mid Sweden University}
  \city{Sundsvall}
  \country{Sweden}}
\email{emin.zerman@miun.se}

\author{Sebastian Knorr}
\orcid{0000-0001-9745-8605}
\affiliation{%
  \institution{HTW Berlin - University of Applied Sciences}
  \city{Berlin}
  \country{Germany}
}

\author{Mårten Sjöström}
\orcid{0000-0003-3751-6089}
\email{marten.sjostrom@miun.se}
\affiliation{%
 \institution{Mid Sweden University}
 \city{Sundsvall}
 \country{Sweden}}

\renewcommand{\shortauthors}{Gond et al.}

\begin{abstract}
  Recent advancements in view synthesis have significantly enhanced immersive experiences across various computer graphics and multimedia applications, including telepresence and entertainment. By enabling the generation of new perspectives from a single input view, view synthesis allows users to better perceive and interact with their environment. However, many state-of-the-art methods, while achieving high visual quality, face limitations in real-time performance, which makes them less suitable for live applications where low latency is critical. In this paper, we present a lightweight, position-aware network designed for real-time view synthesis from a single input image and a target camera pose. The proposed framework consists of a Position Aware Embedding, which efficiently maps positional information from the target pose to generate high dimensional feature maps. These feature maps, along with the input image, are fed into a Rendering Network that merges features from dual encoder branches to resolve both high and low level details, producing a realistic new view of the scene. Experimental results demonstrate that our method achieves superior efficiency and visual quality compared to existing approaches, particularly in handling complex translational movements without explicit geometric operations like warping. This work marks a step toward enabling real-time live and interactive telepresence applications.
\end{abstract}




\begin{teaserfigure}
    \includegraphics[width=\textwidth]{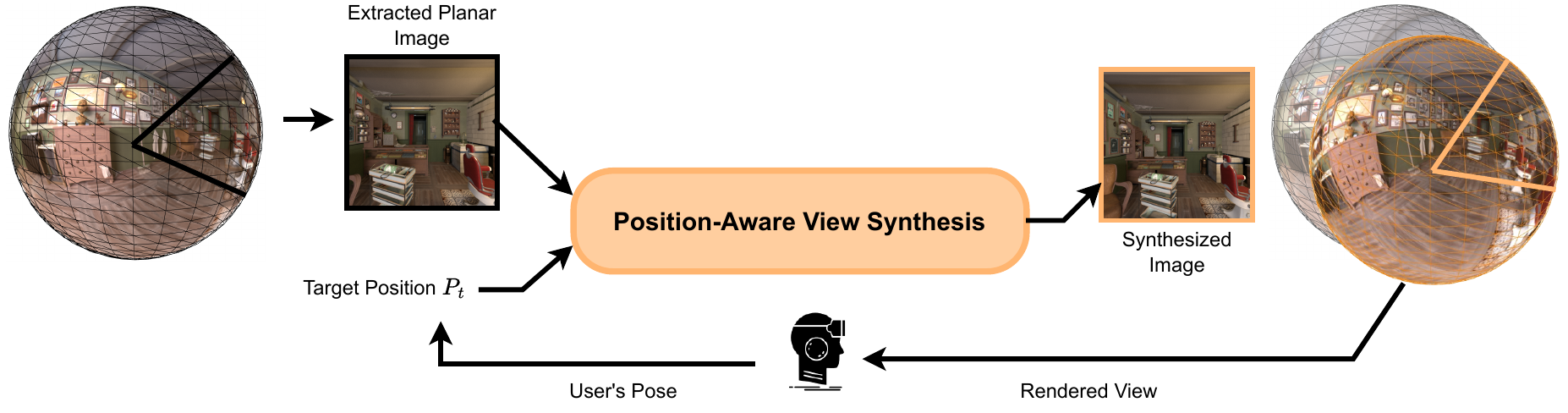}
    \caption{A 360-degree sphere (left) captures the entire environment, from which a highlighted tangent patch is extracted (center-left). This patch is then fed into our position aware view synthesis network (center), which synthesizes a new view (center-right) under a slightly shifted camera pose as per the head movement done by the user. Finally, the synthesized image is mapped back onto the sphere (right) with an offset from the original location, demonstrating the added motion parallax for enhanced depth perception in live applications.}
    \label{fig:teaserImage}
\end{teaserfigure}
\maketitle
\section{Introduction}
\label{sec:intro}
Telepresence systems~\cite{dima2021camera, chandan2021learning, Mose2023VRoxy}, allow users to interact with remote environments in real-time, offering a wide range of applications, particularly in sectors such as heavy machinery, where remote operation from a safe distance is crucial. These systems have the potential to facilitate seamless natural interaction with remote environments. Current real-time telepresence technologies~\cite{chandan2021learning, Mose2023VRoxy} typically rely on head-mounted displays (HMDs), high-speed networks, and consumer or industrial-grade cameras to deliver live video feeds of the environment. While live 2D video feeds provide visual information, they fall  short in offering depth perception~\cite{chandan2021learning}. The lack of depth perception significantly limits situational awareness~\cite{dima2020joint}, which is vital for effective remote operation.

The lack of depth perception that a 2D video feed provides can be leveraged by binocular parallax and motion parallax as depth cues. Recently, SpinVR~\cite{konrad2017spinvr} utilized a stereoscopic virtual reality setup to provide binocular parallax to the users. However, this still lacks motion parallax. When a user moves their viewpoint in space, the relative motion of objects within the scene provides the brain with cues about the depth. Thus, rendering new views at different positions can simulate motion parallax, improving depth perception and enhancing situational awareness in remote operations. For effective depth perception and smooth interaction in real-time, synthesizing new views, while maintaining low latency, becomes essential as it enables both binocular and motion parallaxes.

Synthesizing views from different perspectives requires sufficient information of the scene. While 360-degree cameras are used in some telepresence system~\cite{chandan2021learning}, synthesizing new views from these omnidirectional images (ODIs) as done in works like PanoSynthVR~\cite{PanoSynthVR} introduces additional complexities. The distortion inherent in ODIs and the required high bandwidth for transferring such data make these methods less practical for real-time use. Instead, focusing on planar images captured from the viewpoint of the user offers a more efficient solution. 

\Manu{We consider a telepresence application as the primary use case for our view synthesis method, illustrated in Figure~\ref{fig:teaserImage}. In this setting, an omnidirectional image (ODI) is captured from remote scene. A viewport image is extracted from the ODI based on the current gaze direction of the user, while the relative head movement (i.e., translation) is used to define a target position. The extracted image, target position, and target rotation are then passed to our network, which synthesizes a new view at the intended offset. The resulting image is rendered as the updated viewport, enabling motion parallax without requiring a full 3D reconstruction or stitching it back into the ODI. This patch based rendering pipeline eliminates the distortions which are present in ODIs, reduces transmission bandwidth, and allows for efficient local view synthesis.}

\Manu{By synthesizing views directly from these extracted planar patches, we avoid the distortions inherent in ODIs and reduce the bandwidth needed for transmitting high resolution spherical imagery. This patch based synthesis pipeline supports real-time performance and enables localized motion parallax around the viewpoint of the user. However, such ODI-based workflows inherently embed orientation within the extracted view direction, requiring only translational offsets for rendering. To extend support to general planar RGB inputs such as monocular cameras not embedded in a spherical system, our method also incorporates full 6DoF inputs, including yaw, pitch, and roll. This makes our framework broadly applicable beyond the illustrated telepresence scenario.}

Early \Gond{view synthesis methods~\cite{ahn2013novel, yao2018mvsnet, zhu2019improvedDIBRVR} employed view interpolations~\cite{Chen1993Interpolation} and depth-image-based rendering~\cite{fehn2004DIBR}}, but these suffer from interpolation artifacts, and pixel misalignment due to the use of noisy depth maps. Recent methods like Quark~\cite{Flynn_Quark} and SIMPLI~\cite{solovev2023SIMPLI} employ a dynamic layered depth map (LDM) of the scene estimated through a deep neural network to perform view synthesis. However, these methods still require multi-view input, resulting in requirement of multiple capture devices and high communication bandwidth. Therefore, generating new views from a single input image and a specified camera pose has emerged as a key task in rendering. A variety of approaches, diffusion based 3DiM~\cite{3DiM} and transformer-based architectures like NViST~\cite{jang2024nvist}, are proposed to tackle this problem. These methods leverage large scale datasets to be trained and produce high-quality novel views. \Manu{While these approaches achieve remarkable fidelity, they are either unsuitable for real-time deployment or assume specific scene representations (e.g., dense layers or precomputed geometry). Our work builds on this trajectory by focusing on generalizable, feed forward synthesis that supports both real-time constraints and 6DoF pose inputs.}

\Manu{Real-time rendering is critical for telepresence and AR applications, but most state-of-the-art methods either depend on heavy iterative models (e.g., diffusion~\cite{3DiM, deitke2023objaverse}) or rely on attention-heavy transformer blocks~\cite{jang2024nvist}, both of which impede latency-sensitive deployments.}

In this work, we build upon the growing body of research to address the challenges of computation efficiency and low latency for view synthesis. We propose a method designed to generate high-quality novel views from a single input image, named PVSNet, targeting real-time performance. PVSNet utilizes a position-aware rendering approach that efficiently handles \Manu{camera movements} in the scene, enabling smooth and realistic motion parallax for depth perception without the computational burden of traditional methods.
In summary, our main contributions are as follows:
\begin{itemize}
    \item \Manu{A neural network for real-time view synthesis from a single image and a target pose, supporting both translation only and full 6DoF inputs, making it suitable for telepresence applications.}
    
    \item \Manu{A compact and flexible positional embedding module similar to the one used in NeRF~\cite{nerf}, which is combined with a multi-layer perceptron (MLP), to create a more generalizable framework capable of handling different scenes.}

    \item \Manu{An in-depth evaluation of the proposed method, demonstrating real-time performance with inference times of over 50 FPS at 512$\times$512 resolution, significantly outperforming existing single view methods in speed and image quality.}
\end{itemize}

The remainder of this paper is organized as follows: Section~\ref{sect:background} reviews the state-of-the-art methods aimed at view synthesis. Section~\ref{sect:method_novel_view_synthesis} describes the proposed approach for view synthesis. Section~\ref{sect:experiment} provides a comparative analysis of the results, including quality and speed evaluations. Section~\ref{sect:ablation_study} presents a detailed ablation study to analyze different components of the proposed method. Finally, conclusions are summarized in Section~\ref{sect:conclusion}.

\section{Background}
\label{sect:background}
View synthesis has recently become a pivotal research area in photorealistic immersive telepresence applications, aiming to generate unseen perspectives of a remote scene from a limited set of input images. Traditional approaches, such as structure-from-motion (SfM) and multi-view stereo (MVS), rely on geometric consistency between input views to reconstruct novel viewpoints~\cite{seitz1999photorealistic}. While these methods have shown effectiveness in controlled environments, their performance deteriorates in complex real-world scenarios and often necessitates multiple input views to achieve high-quality results. In light of these limitations, more recent methods focus on synthesizing novel views from sparse or single input images. This section outlines these advancements and their respective contributions to the field of view synthesis based on the number of input images required to perform view synthesis.

\subsection{Multi-View Input}
\label{subsect:background_multi_view_input}
Multi-view input based methods rely on several images or views of a scene to generate novel viewpoints. Recent advancements in multi-view input based methods have centered around leveraging neural implicit representations such as Neural Radiance Fields (NeRF)~\cite{nerf}. A NeRF encodes a scene as a continuous volumetric function, allowing novel view generation through volumetric rendering. Despite its capacity to produce highly photorealistic images, NeRF suffers from computational inefficiencies, particularly in terms of inference time. Several techniques have been proposed to mitigate these limitations, such as hash encoding in InstantNGP~\cite{InstantNGP} and multi-scale representations in Mip-NeRF~\cite{MipNeRF}. While these improvements accelerate rendering, they retain significant reliance on MLPs, which remain computationally expensive.

Plenoxels~\cite{plenoxels}, which leverage spherical harmonics to represent a scene without the need for deep neural networks, have shown promising results when considering their fast optimization time. Though Plenoxels achieve faster optimization than many NeRF-based methods, it still falls short in terms of memory efficiency. A more recent innovation, Gaussian Splatting \cite{kerbl20233d_GSplat}, improves rendering efficiency by employing high-dimensional Gaussian kernels to represent the scene. However, its optimal performance still depends on the availability of multiple input views\Gond{, and the scenes with high frequency details require more kernels, increasing overall memory requirement.}

Light Field Rendering emerges as an alternative of scene regression methods like NeRF~\cite{nerf}, Plenoxels~\cite{plenoxels}, and Gaussian Splatting~\cite{kerbl20233d_GSplat}. Light field (LF) representation encodes both the spatial and angular information of a scene, often described by an HxWxNxM structure, where HxW refers to the spatial resolution and NxM denotes the angular resolution~\cite{levoy1996light}. This representation facilitates tasks like viewpoint manipulation~\cite{Zhou2020}. LF reconstruction methods aim to generate a dense NxM light field from few input views, eliminating the need for complex multi-camera setups. Common LF reconstruction methods ~\cite{Zhou2020,Zhou2021} utilize convolutional neural networks (CNNs) to estimate scene depth followed by a warping operator to reconstruct the full LF. To generate the novel views from the reconstructed LFs, algorithms like dynamically reparameterized LFs~\cite{isaksen2000dynamically} and per-view disk-based blending~\cite{overbeck2018system} have to be employed. However, these methods often have a higher memory footprint, making the process unsuitable for dynamic environments or systems with limited memory.

Despite the effectiveness of LF-based methods, they also face challenges when working in live applications, where real-time performance and memory efficiency are critical. Recent works have introduced more efficient solutions to overcome these issues, such as SIMPLI~\cite{solovev2023SIMPLI}, GNPR~\cite{suhail2022GNPR}, and Quark~\cite{Flynn_Quark}. SIMPLI utilizes a self-improving multiplane-to-layer image approach, enhancing view synthesis by leveraging a deep refinement network for scene representation. However, it still faces challenges in achieving fast inference speeds while maintaining high-quality results. GNPR, on the other hand, uses a generalizable neural framework that extracts patches from neighboring images along the epipolar lines to generate novel views. This method has demonstrated impressive quality, but like SIMPLI, it struggles with high computational demands and inefficiencies during inference. In contrast, Quark introduces a novel layered depth map (LDM) approach that combines efficiency and quality by performing a multi-scale iterative render-and-refine process. This method achieves real-time, high-resolution rendering, surpassing SIMPLI and GNPR in terms of both speed and quality. Despite its impressive performance, Quark remains limited to multi-view inputs, restricting its applicability in scenarios where only a single input image is available. 

Our proposed method, PVSNet described in Section~\ref{sect:method_novel_view_synthesis}, goes a step further by offering a more efficient view synthesis approach, utilizing only a single input image. By removing the dependency on multiple views, PVSNet not only reduces the computational overhead, but also extends the applicability of view synthesis methods to real-time environments. In addition, the method ensures that high-quality novel views are generated while maintaining efficiency, making it suitable for resource-constrained applications. \Manu{Unlike methods that rely on fixed view grids or dense multi-camera capture, PVSNet operates on arbitrary camera positions and orientations, making it suitable for unconstrained, real-time scenarios. The ability to embed full 6DoF poses directly into the synthesis pipeline allows our method to generalize across both panoramic and planar input modalities without requiring view-specific encodings or ray models.}

\subsection{Single-View Input}
\label{subsect:background_single_view_input}
Single-view input methods aim to synthesize novel views from just one image, primarily addressing the scarcity of limited input data. Many of these techniques employ deep learning models to infer depth or scene structure from a single image. For instance, PixelNeRF \cite{yu2021_pixelnerf} and VisionNeRF \cite{lin2023_visionnerf} generalize NeRF to new scenes by conditioning on image features at large scale. Although the PixelNeRF and VisionNeRF show promising results, they face challenges such as scale ambiguity and large baseline camera movements, which degrade rendering quality in more extreme viewpoints.

\begin{figure*}
    \centering
    \includegraphics[width=0.95\textwidth]{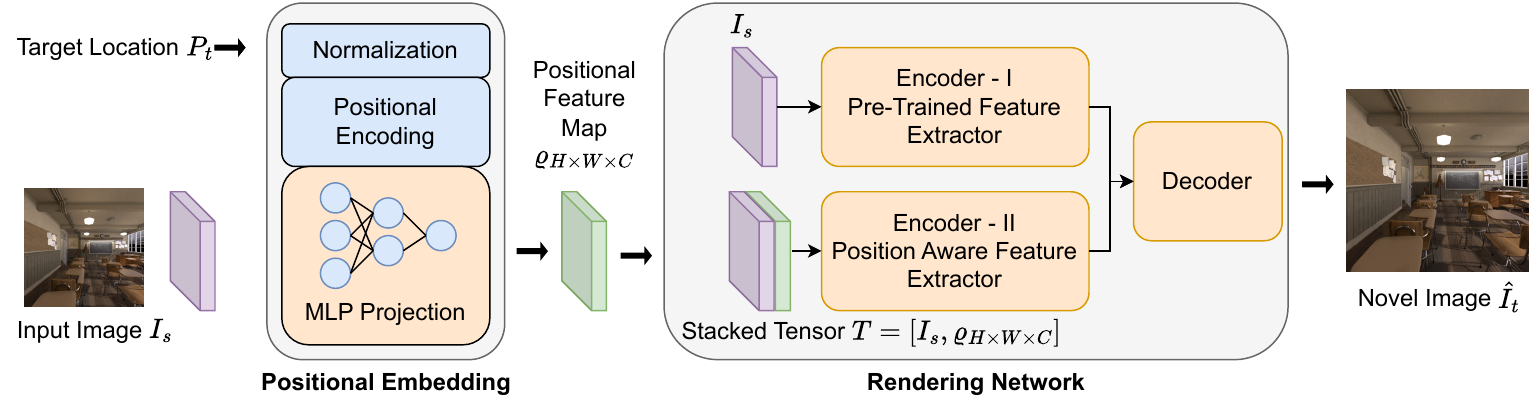}
    \caption{Pipeline of view synthesis using \textbf{PVSNet} which takes single input image \(I_s\) \& target location \(P_t\), and produces the image \(\hat{I}_t\). The Positional embedding branch takes \(P_t\) and creates a higher dimensional projection of positional feature \(\mathbf{\varrho}_{H \times W \times C}\), which then along with input image \(I_s\) is passed to rendering network responsible for producing output image \(\hat{I}_t\).}
    \vspace{-0.3cm}
    \label{fig:PLFNet_Plus_Block}
\end{figure*}

To mitigate issues like scale ambiguity and substantial viewpoint changes, diffusion-based approaches have emerged. For example, 3DiM~\cite{3DiM} employs pose guided diffusion to improve performance, particularly in scenarios with large translational movement. Fine-tuning diffusion models on large-scale synthetic datasets~\cite{deitke2023objaverse} has also shown potential~\cite{liu_image3d, liu_mesh3d} in handling the scale ambiguity, although these methods are hindered by slow inference times, and highest memory footprint compared to all other methods discussed so far.

\Manu{Other single-view input methods such as 3D Photo~\cite{shih20203d} and SLIDE~\cite{jampani2021slide} focus on inferring views through learned layered depth inpainting. These methods can yield visually compelling results, especially for large baselines, but often require RGB+D input~\cite{shih20203d} or rely on multi-stage pipelines with monocular depth estimation, layering, and hole-filling~\cite{jampani2021slide}, making them unsuitable for real-time applications.}

As an alternative to diffusion-based frameworks, multiplane image (MPI) representations approximate a scene using a series of parallel planes at fixed depths, each containing a color and alpha map. This representation allows view synthesis based on a fixed~\cite{Zhou2018MPI,Flynn_2019_CVPR,tucker2020single,Li2020} or variable~\cite{Li2020} number of planes. While MPI-based methods achieve superior visual quality, their computational demands remain high, especially during MPI layer prediction. 

More recent innovations, such as NViST~\cite{jang2024nvist}, incorporate vision transformers (ViTs)~\cite{ViT} into view synthesis architectures, demonstrating strong generalization capabilities in real-world settings. By leveraging masked autoencoders (MAEs) and adaptive layer normalization conditioned on camera parameters, NViST delivers high-quality results, even for out-of-distribution views. Despite these strengths, the reliance of NViST on a ViT backend and patch-wise image processing leads to slower inference at higher resolutions. Meanwhile, large-scale GANs~\cite{koh2023simple_se3d} offer rapid rendering and handle large baselines effectively at high resolution from a single ODI, but it also incurs significant memory overhead, limiting the practical usage in resource-constrained settings. \Manu{Another relevant line of work is MegaParallax~\cite{bertel2019megaparallax}, which aims to produce motion parallax in ODI using layered mesh generated from monocular depth maps and image warping. While it succeeds in synthesizing novel viewpoints from a single ODI, the pipeline involves explicit depth estimation, foreground segmentation, and layered rendering, which introduce computational overhead and make real-time operation impossible.}

In summary, despite the advances, single-view methods frequently grapple with achieving both real-time performance and high-quality novel view generation in live applications. Addressing this balance between efficiency and fidelity remains a critical goal for immersive telepresence and related applications. The limitations identified here, which includes scale ambiguity, heavy computational demands, and memory constraints, inform the design of our proposed approach PVSNet (described in Section~\ref{sect:method_novel_view_synthesis}), which aims to reconcile the trade-offs these methods face.



\section{Method: Position-Aware View-Synthesis}
\label{sect:method_novel_view_synthesis}
PVSNet is a two stage model which, given the source image $I_s$ and the target position $P_t$, aims to synthesize the novel image $I_t$ at the corresponding target viewpoint. Our approach consists of two key stages as shown in Figure~\ref{fig:PLFNet_Plus_Block}. First, the target position is embedded into a high-dimensional space through positional embedding which is discussed in Section~\ref{sect:method_position_embedding}, and second, new views are rendered from these positions, which is detailed in Section~\ref{sect:rendering_network_plfnet_plus}.

\subsection{Positional Embedding}
\label{sect:method_position_embedding}
\Manu{We aim to synthesize views from a single input image $I_s$ and a target pose $P_t \in \mathbb{R}^6$, defined by 3D translation $(x, y, z)$ and rotation $(\theta, \phi, \psi)$ representing yaw, pitch, and roll, respectively. While recent approaches such as \emph{Cameras as Rays}~\cite{zhang2024cameras} propose learning patch wise ray bundles for pose estimation, these are designed for multi-view image setups and rely on ray supervision and diffusion sampling. In contrast, our goal is real-time view synthesis from a single image, where camera pose is already known. Thus, we adopt a compact 6D vector representation and learn a high-dimensional embedding to support feed-forward synthesis at high frame rates.}

\Manu{Although direct regression using $P_t$ may suffice for small viewpoint changes, it becomes insufficient for larger baselines. To better capture pose dependent variations,} we employ a positional embedding strategy that combines positional encoding, inspired by NeRF~\cite{nerf}, along with our proposed MLP module for enhanced representation of the target coordinate $P_t$.

\textbf{Coordinate Normalization:}
First, the target coordinates \Manu{denoted by $P_t(x,y,z,\theta, \phi, \psi)$} are normalized into \Manu{$\bar{P}_t(\bar{x}, \bar{y}, \bar{z}, \bar{\theta}, \bar{\phi}, \bar{\psi})$} to ensure consistency across varying ranges of viewpoints \Manu{in scenes and training samples}:
\begin{equation}
    \label{eqn:normalized_coordinates}
    \Manu{\bar{P}_t = \frac{P_t - P_{\min}}{P_{\max} - P_{\min}},}
\end{equation}
\Manu{where $P_{\min}, P_{\max} \in \mathbb{R}^6$ denote the minimum and maximum values across the training set for each component of $P_t$. This ensures all elements of the normalized pose vector $\bar{P}_t \in [0, 1]^6$.} This normalization step ensures that the coordinates lie within a fixed range, which is crucial for subsequent encoding.

\textbf{Positional Encoding:}
To project the normalized coordinates $\bar{P}_t$ into a higher-dimensional space, we apply positional encoding. This involves concatenating multiple sine and cosine functions of varying frequencies

\begin{equation}\label{eqn:positional_encoding}
\gamma(\bar{P}_t) = \left( \dots, \cos{\left( 2\pi \sigma^{\frac{j}{m}} \bar{P}_t \right)}, \sin{\left( 2\pi \sigma^{\frac{j}{m}} \bar{P}_t \right)}, \dots \right) \in \mathbb{R}^{1 \times d_1},
\end{equation}
where \Manu{\(\bar{P}_t = (\bar{x}, \bar{y}, \bar{z},\bar{\theta}, \bar{\phi}, \bar{\psi})\)} is the normalized \Manu{6D} coordinate vector, $m$ represents the encoding depth, and \(j \in \{0, \dots, m-1\}\). The parameter \(\sigma\) is a scaling constant based on the domain of \(\bar{P}_t\). For our experiments, we use $m = 32$, resulting in an output size of \Manu{$d_1 =2 \cdot m \cdot 6 = 384$}  dimensions, which provides enhanced positional awareness (see Table~\ref{tab_ablation_study_positional_embedding} for ablation study results).

\textbf{Learning Based Reprojection:}
Once the positional encoding \(\gamma(\bar{P}_t)\) has been computed, we further enhance the representation by passing it through an MLP. This MLP maps the given encoded vector into a much higher-dimensional space, which has proven to work well with scene regression task as shown in NeRF~\cite{nerf}. This process allows for a more expressive and flexible feature space which,  unlike NeRF, can be adapted to different scenes since we only consider the coordinates in this representation and not the scene properties at this point. Our approach is similar to the adaptive layer normalization techniques used in NViST~\cite{jang2024nvist}.

Let \(\gamma(\bar{P}_t) \in \mathbb{R}^{1 \times d_1}\) be the output of the positional encoding, where \(d_1\) is the dimensionality of the encoding. The MLP projects this vector into a high-dimensional space \(\mathbb{R}^{1 \times d_2}\), where $d_2$ is chosen based on the desired resolution: 
\begin{equation}\label{eqn:mlp_projection_variable}
    \mathbf{\rho} = \text{MLP}\left( \gamma(\bar{P}_t) \right) \in \mathbb{R}^{1 \times d_2}.
\end{equation}
Then, the vector \(\mathbf{\rho}\) is reshaped into a feature map $\mathbf{\varrho}_{h \times w \times c}$ of size $h \times w \times c$, where \Gond{$d_2 = h w$, w:h has the same aspect ratio as the original image,} and $c=1$. This reshaping operation can be expressed using the inverse vectorization notation as
\begin{equation}\label{eqn:reshape_variable}
    \mathbf{\varrho}_{h \times w \times c} = \text{vec}^{-1}(\mathbf{\rho}) \in \mathbb{R}^{h \times w \times c}.
\end{equation}
Finally, the feature map \(\mathbf{\varrho}_{h \times w \times c}\) is upsampled \Manu{using nearest neighbor method} to match the spatial resolution of the input image, denoted by $H \times W$, thus we set $C=1$. This upsampling is performed using a scaling factor \(\alpha = \frac{H}{h} = \frac{W}{w}\), as follows
\begin{equation}\label{eqn:upsample_variable}
    \mathbf{\varrho}_{H \times W \times C} = \text{Upsample}\left( \mathbf{\varrho}_{h \times w \times 1}, \alpha \right) \in \mathbb{R}^{H \times W \times C}.
\end{equation}
The final upsampled representation \(\mathbf{\varrho}_{H \times W \times C}\) retains the positional information of the target coordinate and is subsequently used by the rendering network to generate the novel view.

\subsection{Rendering Network}
\label{sect:rendering_network_plfnet_plus}
The role of the rendering network in PVSNet is to synthesize a target image \( I_t \) from the input image \( I_s \) and the positional embedding \( \mathbf{\varrho}_{H \times W \times C} \), obtained from the positional embedding stage. Our network architecture is inspired by LFSphereNet~\cite{gond2023lfspherenet} and consists of two distinct encoder branches followed by a decoder network. These branches work in tandem to extract and merge spatial and positional features, ultimately generating high quality novel views.

The rendering network operates as follows: the first encoder processes the input image \( I_s \) to extract image-based features, while the second encoder processes both the input image \( I_s \) and the positional feature map \( \mathbf{\varrho}_{H \times W \times C} \) to capture position-dependent information. These features are subsequently merged in the decoder, which refines and upsamples the combined feature maps to generate the final output \( \hat{I}_t \), a novel view at the target position.

\subsubsection{Encoder I - Pre-trained Image Feature Extractor}
\label{sect:encoder_1}
The first encoder, \( E_1 \), is a pre-trained image feature extractor that processes the input image \( I_s \). Following the approach used in ResNet-152~\cite{he2016deep}, we utilize the initial layers of ResNet to extract lower-level feature maps that are rich in texture and local detail. Specifically, \( E_1 \) extracts a set of feature maps \( F_1 \in \mathbb{R}^{h_1 \times w_1 \times 256} \), where \( h_1 \) and \( w_1 \) represent the spatial dimensions of the downsampled feature maps. These feature maps are then expanded to \( F_1' \in \mathbb{R}^{h_1 \times w_1 \times 512} \) through additional convolution layers to increase the depth of the feature space which can be described as
\begin{equation}
    F_1^\prime = \text{Conv}\left( E_1(I_s) \right) \in \mathbb{R}^{h_1 \times w_1 \times 512}.
\end{equation}
The extracted features \( F_1' \) capture essential image content but lack positional awareness, which is handled by the second encoder.

\subsubsection{Encoder II - Position-Aware Feature Extractor}
\label{sect:encoder_2}
The second encoder, \( E_2 \), integrates both the input image \( I_s \) and the positional embedding \( \mathbf{\varrho}_{H \times W \times C} \) to adaptively capture position-specific features. Unlike the first encoder, which only processes the image, \( E_2 \) focuses on refining the feature maps based on the target position. This is done by capturing the high‐frequency variations in geometry and appearance much more effectively.

We first concatenate the input image \( I_s \) and the positional embedding \( \mathbf{\varrho}_{H \times W  \times C} \) along the channel dimension, resulting in a stacked input tensor
\begin{equation}
    T = \left[ I_s, \mathbf{\varrho}_{H \times W} \right] \in \mathbb{R}^{H \times W \times (3+C)},
\end{equation}
where \( C \) is the number of channels in \( \varrho \). In our case, \( C =  1\).

This stacked tensor \( T \) is passed through a series of convolutional layers that preserve the spatial dimensions, ensuring pixel-level positional information is retained throughout the encoding process, as demonstrated by LFSphereNet~\cite{gond2023lfspherenet}. \Gond{In the last two layers, Max Pooling is employed to downsample the feature maps to match the spatial dimension of feature map \(F_1'\).} The encoder produces a feature map \( F_2 \in \mathbb{R}^{h \times w \times d_3} \), where \( d_3 \) is the number of feature channels. Additionally, skip connections are generated from intermediate layers for later use in the decoder.

Formally, this can be expressed as
\begin{equation}
    F_2 = E_2\left( \left[ I_s, \mathbf{\varrho}_{H \times W} \right] \right) \in \mathbb{R}^{h \times w \times d_3}.
\end{equation}

The skip connections, denoted as \( S_k \), are collected from intermediate layers, where \( k \) indexes the corresponding layer, and are used to enhance spatial detail during the decoding stage.

\subsubsection{View Decoder}
\label{sect:view_decoder}
The decoder, \( D \), is responsible for merging the feature maps from both encoder branches and reconstructing the target image \( \hat{I_t} \) at the desired target position. The concatenated feature maps \( F_1' \) and \( F_2 \) are first combined along the channel dimension
\begin{equation}
    F_{concat} = \left[ F_1', F_2 \right] \in \mathbb{R}^{h \times w \times (512 + d_3)}.
\end{equation}

This concatenated feature map is then passed through a series of transposed convolution layers to progressively upsample the spatial resolution, ultimately matching the resolution of the input image. The decoder uses a combination of transposed convolutions and bilinear upsampling, preserving spatial coherence in the generated image. Skip connections \( S_k \) from the second encoder are concatenated into the upsampled feature maps at various stages to retain high-frequency details. The skip connections ensure that fine-grained positional and image details are retained, resulting in more consistent and accurate novel views. The overall process can be expressed as
\begin{equation}
    \hat{I}_t = D\left( F_{concat}, S_1, S_2, \dots, S_n \right) \in \mathbb{R}^{H \times W \times 3}.
\end{equation}
The final output of the decoder is the synthesized target image \( \hat{I}_t \), with spatial dimensions \( H \times W \).

\subsection{Training Loss}
\label{sect:method_nvs_training_loss}
To optimize our PVSNet, we combine four loss functions: L1 loss, Multi-Scale Structural Similarity (MS-SSIM)~\cite{wang2003multiscale} loss, Focal Frequency Loss (FFL)~\cite{jiang2021focal_loss} \Manu{and perceptual loss~\cite{chen2017photographic}}. Each of these losses contributes to different aspects of the synthesis process, improving both spatial and frequency domain accuracy.

\textbf{L1 Loss and MS-SSIM Loss} are combined to capture fine-grained details, especially when dealing with larger baselines. The L1 loss minimizes pixel-wise differences between the predicted novel view image $\hat{I}_t$ and the ground truth image $I_t$, ensuring accurate reconstruction of intensity values. On the other hand, MS-SSIM is a perceptually motivated metric that focuses on preserving structural information and texture at multiple scales, correlating better with human visual perception, as demonstrated in image restoration tasks~\cite{zhao2016loss}.

\textbf{Focal Frequency Loss (FFL)} is utilized to enhance the quality of the synthesized views, particularly in the frequency domain. We incorporate Focal Frequency Loss (FFL)~\cite{jiang2021focal_loss} to address spectral bias in neural networks by focusing on the frequency components that are hardest to synthesize. This helps the model to improve the reconstruction of high-frequency details, which are crucial for generating realistic novel views.

The FFL is defined as:
\begin{equation}
    L_{\text{FFL}} = \frac{1}{M N} \sum_{u=0}^{M-1} \sum_{v=0}^{N-1} w(u,v) \left| F_r(u, v) - F_f(u, v) \right|^2,
\end{equation}
where $F_r(u, v)$ and $F_f(u, v)$ represent the real and predicted frequency components at position $(u, v)$ in the Fourier domain, and $w(u,v)$ is a dynamic weight that emphasizes the hardest-to-synthesize frequencies.

\Manu{\textbf{Perceptual Loss} is used to improve the perceptual quality of the synthesized views by comparing high-level feature activations between the predicted and ground truth images. Inspired by~\cite{chen2017photographic}, we extract intermediate feature maps from a pretrained VGG-19 network and minimize the L1 difference between corresponding layers. These layers represent increasing levels of abstraction from edges to object parts and thus guide the synthesis model to preserve both fine grained detail and global structure. Specifically, we define the perceptual loss as:}

\begin{equation}
\Manu{L_{\text{VGG}} = \sum_{l} \lambda_l \cdot | \Phi_l(\hat{I}_t) - \Phi_l(I_t) |_1,}
\end{equation}
\Manu{where $\Phi_l$ denotes the activation from layer $l$ of the VGG-19 network, and $\lambda_l$ is a weighting factor inversely proportional to the number of elements in that layer. This perceptual supervision helps produce sharper, more semantically coherent outputs, particularly in areas where pixel-wise losses are insufficient to capture visual fidelity.}

\textbf{The total loss} function $L_{\text{total}}$ for PVSNet is then defined as:
\begin{equation}
    L_{\text{total}} = \alpha \cdot L_{\text{L1}} + (1 - \alpha) \cdot L_{\text{MS-SSIM}} + \beta \cdot L_{\text{FFL}} \Manu{+ L_{\text{VGG}},}
\end{equation}
where $\alpha$ is the weighting factor that balances the contribution of L1 and MS-SSIM losses and $\beta$ is a weight factor that balances the contribution of the FFL relative to the L1 and MS-SSIM losses. This combined loss function enables PVSNet to generate sharper and more realistic novel views by optimizing for both spatial accuracy and high-frequency details.

\begin{figure*}
    \centering
    \includegraphics[width=0.95\textwidth]{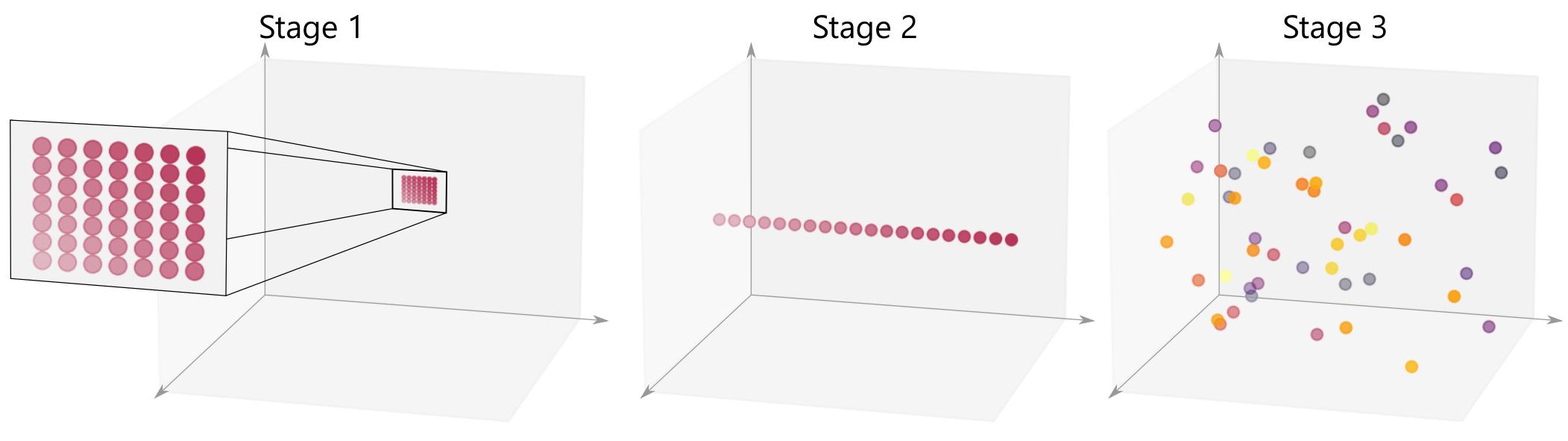}
    \caption{Conceptual view of camera placement of each stage of training. Stage 1 involves small baseline and a fixed grid of cameras, stage 2 focuses on larger baseline but still keep the discrete placement of cameras on a fixed grid. Finally, stage 3 trains the network on large baseline with random camera placements in a cubical volume. The gray volume roughly describes a 20 $cm^3$ volume. Bright colors on the rightmost figure are closer to us and darker ones are away.}
    \vspace{-0.3cm}
    \label{fig:multi_stage_training}
\end{figure*}

\subsection{Implementation}
\label{sect:method_implementation_nvs}
The input of PVSNet is a single RGB image \(I_s\) and the target position \Gond{\(P_t\), for the positional embedding output \(\mathbf{\varrho}_{H \times W \times C}\) }we set $C=1$ since we use a single channel positional map.
PVSNet was implemented in \Gond{PyTorch\footnote{\label{code_note}
Code and checkpoint for our model will be made available upon acceptance of this paper at \url{https://github.com/Realistic3D-MIUN/PVSNet}}}, and trained with the following hyperparameters: batch size of 24, learning rate of 0.003, adam optimizer, MS-SSIM, L1 \Gond{and FFL loss.} After each 30 epochs, the learning rate was decreased by a factor of 0.2. The network was trained for 150 epochs on 2 GPUs (Nvidia A40)  within a computing cluster with Intel Xeon Gold 6338 CPUs. The training duration was 48 hrs. Due to limitation of training data when using our synthetic dataset we employ multi stage training strategies to reach satisfactory results in the training process.

\subsubsection{Multi Stage Training}
\label{sect:method_implementation_nvs_multi_stage_training}
Given the scarcity of large-scale datasets for view synthesis with wide baselines, we adopt a three-stage training approach shown in Figure~\ref{fig:multi_stage_training} to progressively enhance the network's ability to capture image features and handle positional shifts.

\textbf{Stage 1} serves as a preconditioning step, training the network on a large-scale light field dataset~\cite{Srinivasan2017}. While the baseline between the views in each light field is relatively narrow, the large scale of dataset enables the network to learn both high and low level image features, implicitly understanding depth. In this stage, the network is conditioned to synthesize novel views within a constrained light field volume. Once the network demonstrates proficiency in this task, we transition to the next stage.

\textbf{Stage 2} introduces a dataset with a significantly larger baseline: the Spherical Light Field Database (SLFDB)~\cite{zerman2024_slfdb}, which consists of 20 spherical light fields captured with 60 angular dimensions. From these omnidirectional images (ODIs), we extract planar patches and fine-tune the network with a lower learning rate. This step enables the network to learn to handle wider baseline movements while maintaining positional accuracy. Following this, the network is prepared for the final stage of training, which focuses on the joint optimization of the MLP and rendering network with more random positions.

\textbf{Stage 3} involves training the network on our synthetic dataset (discussed in Section~\ref{sect:experiments_nvs_datasets}) containing random camera viewpoints within a \(N \times N \times N\) cubical volume. The increase in the spatial search space necessitates the use of the positional embedding module for accurate view synthesis. In this stage, the network is fine-tuned to handle more complex camera translations and larger scene variations, fully utilizing the positional embedding to enhance rendering precision.

\begin{figure}[]
\centering
\begin{tabular}{cc}
 Classroom &
 Italian Flat \\
 \includegraphics[width=1in]{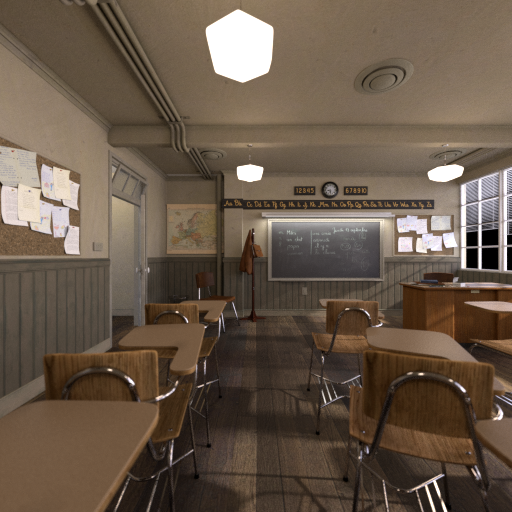} &
 \includegraphics[width=1in]{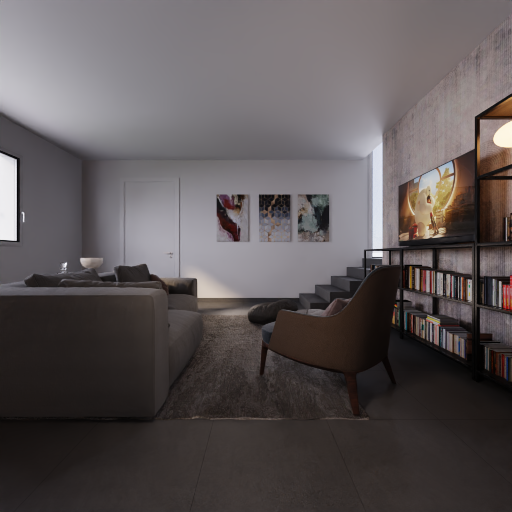} \\
 Lone Monk &
 Barbershop \\
 \includegraphics[width=1in]{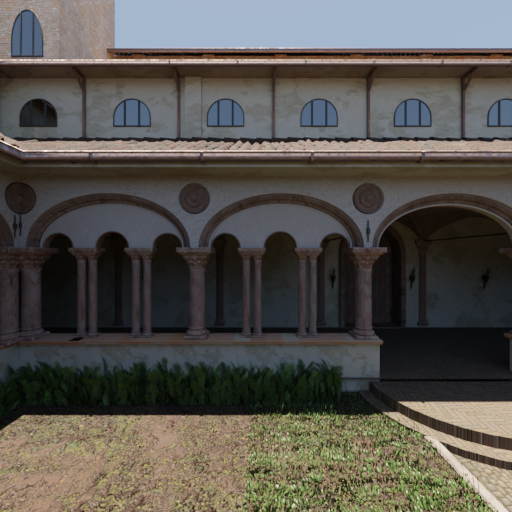} &
 \includegraphics[width=1in]{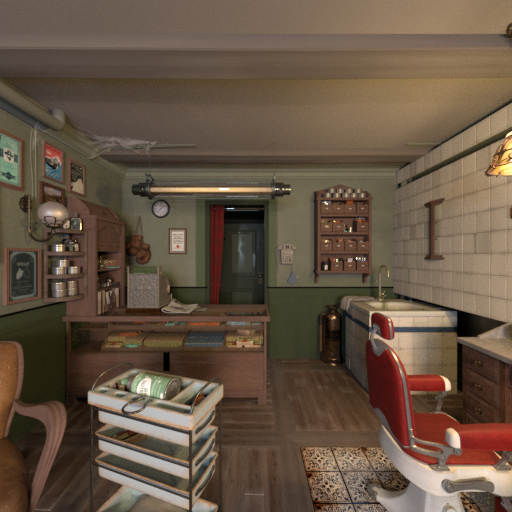} \\
\end{tabular}
\caption{Some example images taken from each unique scene from the blender dataset.}
\label{fig:blender_dataset}
\end{figure}

\section{Experimental Evaluation}
\label{sect:experiment}
In this section, we present a comprehensive evaluation of our proposed PVSNet architecture. \Gond{We begin by assessing the performance of PVSNet for view synthesis on datasets with varying baselines and complexities. Following this, we demonstrate the capability of PVSNet for small baseline light field reconstruction and view synthesis within LF baseline from a single input image.} The experiments for view synthesis are detailed in Section~\ref{sect:experiments_novel_view_synthesis}, while light field reconstruction results are discussed in Section~\ref{sect:experiments_light_field_reconstruction}.


\begin{table*}[t]
  \centering
  \caption{View Synthesis Results (best values in bold): Quantitative comparison of view synthesis methods on Blender (256x256 and 512x512) and COCO (256x256 and 512x512) datasets. PVSNet (Ours) achieves the best overall results on the Blender dataset across most metrics, significantly outperforming previous methods. On the COCO dataset, PVSNet demonstrates strong performance, achieving the second-best scores, particularly excelling in SSIM and MS-SSIM, closely trailing AdaMPI. These results highlight PVSNet's capability to generate high-quality novel views with competitive accuracy, especially in complex real-world settings like COCO.}
  \label{tab_nvs}
  \begin{tabular}{l*{11}{c}r}
     \hline
    Dataset     & Method        & PSNR $\uparrow$   & SSIM $\uparrow$   & MS-SSIM $\uparrow$    & VIFP $\uparrow$   & DISTS $\downarrow$      & LPIPS $\downarrow$\\
    \hline
                & SinMPI~\cite{pu2023sinmpi}        & 17.3295           & 0.4193            & 0.7064                & 0.4911            & 0.2794                 & 0.4097\\
    Blender     & TMPI~\cite{khan2023tiled_tmpi}           & 19.9507            & 0.7160            & 0.8264                & 0.4822           & 0.1458                 & 0.1471\\
    256x256     & AdaMPI~\cite{han2022single_AdaMPI}        & 24.9252           & 0.8092            & 0.9145                & 0.6720            & 0.0680        & 0.0455\\
                & PVSNet (Ours)        & \textbf{31.2258}  & \textbf{0.9386}   & \textbf{0.9793}       & \textbf{0.8702}   & \textbf{0.0613}                 & \textbf{0.0275}\\
                & PVSNet-Lite (Ours)                       & 28.7894  & 0.8902   & 0.9577       & 0.7655   & 0.0942                 & 0.0566\\
                
    \hline
                & SinMPI~\cite{pu2023sinmpi}        & 18.3441           & 0.4599            & 0.6885                & 0.3621            & 0.2251                 & 0.3966\\
    Blender     & TMPI~\cite{khan2023tiled_tmpi}           & 20.0126           & 0.7071            & 0.8070                & 0.3605            & 0.1268                 & 0.1640\\
    512x512     & AdaMPI~\cite{han2022single_AdaMPI}        & 23.1416           & 0.7510            & 0.8566                & 0.4353            & 0.0751         & 0.0770\\
                & PVSNet (Ours)        & \textbf{30.0085}  & \textbf{0.9254}   & \textbf{0.9733}       & \textbf{0.8408}   & \textbf{0.0733}                  & \textbf{0.0329}\\
                & PVSNet-Lite (Ours)                       & 25.8741  & 0.8135   & 0.9100       & 0.6063   & 0.0852                 & 0.0778\\
    \hline          
                & SinMPI~\cite{pu2023sinmpi}                     & 14.8544           & 0.4669            & 0.6484                & 0.3521            & 0.2940                 & 0.5111\\
    COCO~\cite{caesar2018coco}  & TMPI~\cite{khan2023tiled_tmpi} & 17.6991            & 0.5822            & 0.7411                & 0.4032            & 0.2161                 & 0.2490\\
    256x256     & AdaMPI~\cite{han2022single_AdaMPI}              & 22.1027            & 0.8093            & 0.9241                & 0.7180            & 0.1152                 & 0.0849\\
                & PVSNet (Ours)              & \textbf{24.0435}   & \textbf{0.8462}    & \textbf{0.9445}   & \textbf{0.7518}       & \textbf{0.0825}   & \textbf{0.0600}\\
                & PVSNet-Lite (Ours)                       & 21.3391  & 0.7555   & 0.9122       & 0.6195   & 0.1147                 & 0.0915\\
    \hline          
                & SinMPI~\cite{pu2023sinmpi}              & 15.1025            & 0.4561            & 0.6500                & 0.3149            & 0.2250                 & 0.3555\\
    COCO~\cite{caesar2018coco} & TMPI~\cite{khan2023tiled_tmpi} & 18.2453            & 0.6290            & 0.7851                & 0.3003            & 0.1750         & 0.2711\\
    512x512     & AdaMPI~\cite{han2022single_AdaMPI}              & 21.3099            & 0.7356            & 0.8792               & 0.5099            & 0.1196                 & 0.1673\\
                & PVSNet (Ours)              & \textbf{22.8850}            & \textbf{0.7764}            & \textbf{0.9015}                & \textbf{0.6095}            & \textbf{0.1070}                 & \textbf{0.1437}\\
                & PVSNet-Lite (Ours)                       & 19.8910  & 0.6200   & 0.7842       & 0.4613   & 0.1364                 & 0.1894\\
    \hline  
\end{tabular}
\end{table*}

\begin{table}[t]
  \centering
  \caption{Inference time (in seconds) to render the novel view for different resolutions on RTX 2070 Super or and frames per second (in seconds / fps). Best values in bold. The fps values are rounded down for simplicity.}
  \label{tab_nvs_runtime}
  \begin{tabular}{l*{12}{c}r}
     \hline
    Method     & 256x256    & 512x512   \\
    \hline
    SinMPI~\cite{pu2023sinmpi}          & 0.0752 / 13     & 0.2368 / 4    \\
    TMPI~\cite{khan2023tiled_tmpi}      & 0.0910 /11     & 0.2640 / 4   \\
    AdaMPI~\cite{han2022single_AdaMPI}  & 0.2261 / 4     & 0.3254 / 3   \\
    PVSNet (Ours)                              & 0.0161 / 62     & 0.0181 / 55    \\
    PVSNet-Lite (Ours)                             & \textbf{0.0081 / 123}     & \textbf{0.0096 / 104}    \\
    \hline
\end{tabular}
\end{table}

\subsection{View Synthesis}
\label{sect:experiments_novel_view_synthesis}
\subsubsection{Datasets} 
\label{sect:experiments_nvs_datasets}

To evaluate the performance of PVSNet on view synthesis, we use two datasets: a custom Blender dataset tailored for this task and the COCO dataset~\cite{caesar2018coco} for more diverse scenes to demonstrate broader generalizability.

\textbf{Blender dataset} was created by selecting four freely available virtual scenes from the Blender demo website---namely \textit{Lone Monk, Classroom, Barbershop, and Italian Flat} as shown in Figure~\ref{fig:blender_dataset}. For each scene, we first positioned the camera at various initial locations and then applied random camera movements constrained within a \(20\,\text{cm}\) cubical volume around each initial position. These random movements allowed us to capture multiple images per scene, all within the specified \(20\,\text{cm}^3\) region. The resulting set of images from these scenes constitutes our custom Blender dataset\footnote{\label{dataset_note}
Dataset url will be made available upon acceptance of this paper}.

\textbf{COCO dataset} was adopted with a similar approach to AdaMPI \cite{han2022single_AdaMPI} for generating training data, which involved utilizing depth-image-based rendering (DIBR) with hole filling. During evaluation, we take an input image and synthesize a novel image at a specified target location. The synthesized image is then treated as input to generate an image back at the original location. This process enables us to compare the reconstructed image with the ground truth (the original input image), thereby allowing for an assessment of rendering quality.
Both datasets are evaluated at resolutions of \(256 \times 256\) and \(512 \times 512\).

\begin{figure}
    \centering
    \includegraphics[width=0.45\textwidth]{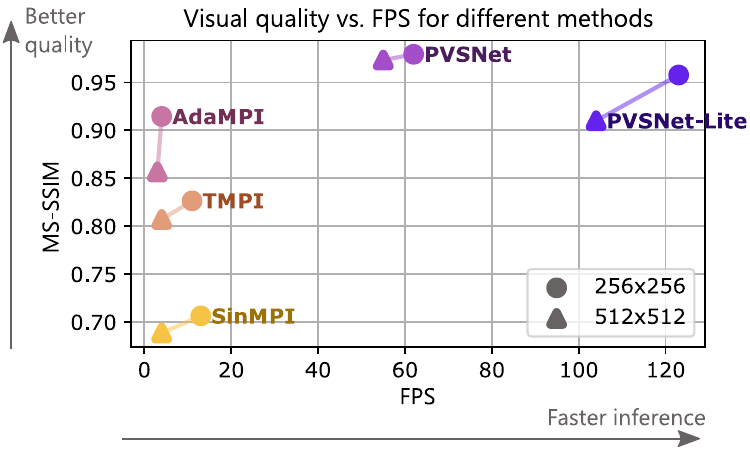}
    \caption{Quality vs FPS rate of different methods against different resolution for Blender dataset.}
    \label{fig:Quality_vs_FPS}
\end{figure}

\begin{figure*}
    \centering
    \includegraphics[width=0.95\textwidth]{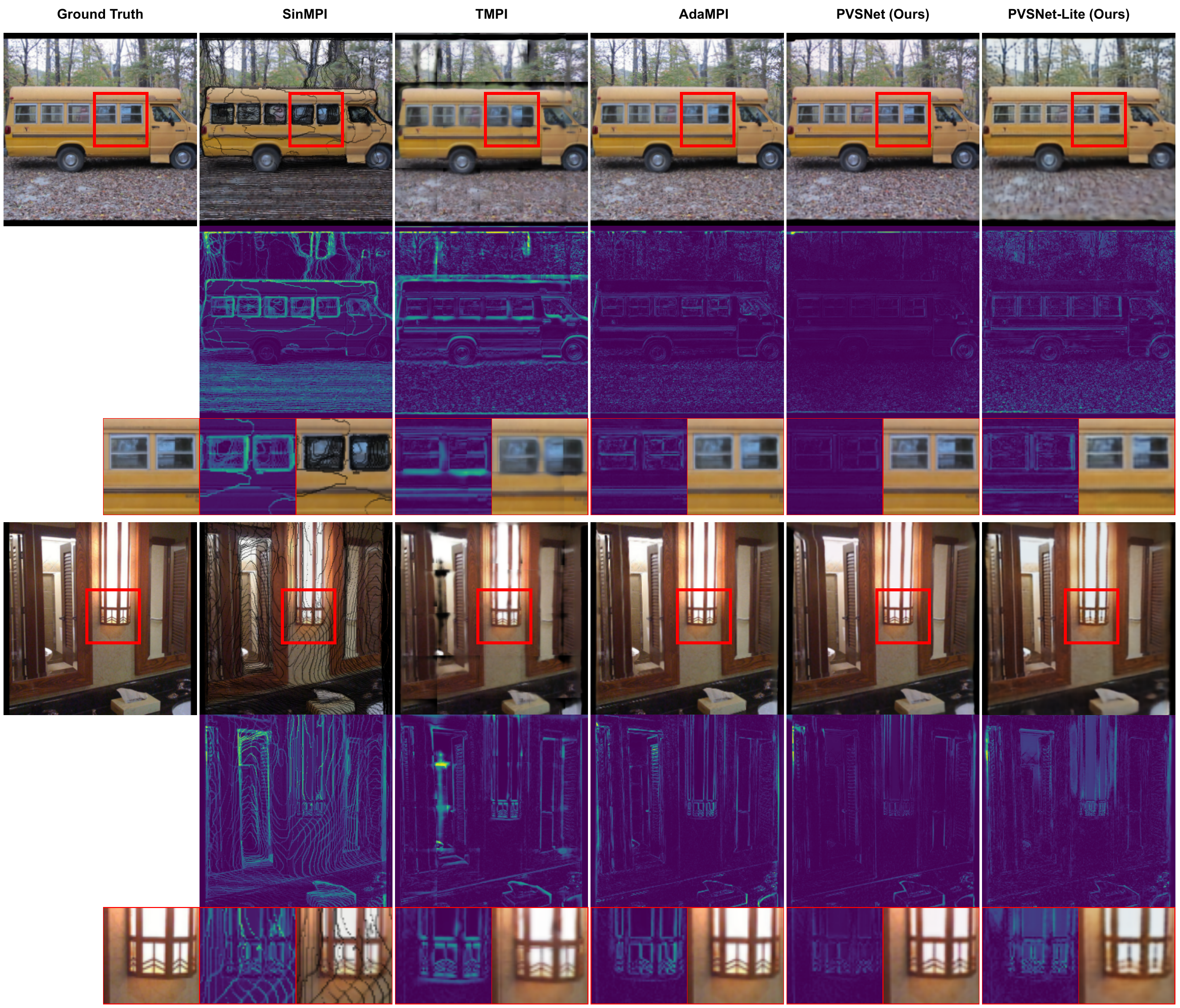}
    \caption{Comparison of view synthesis results on COCO dataset with error maps for SinMPI~\cite{pu2023sinmpi}, TMPI~\cite{khan2023tiled_tmpi}, AdaMPI~\cite{han2022single_AdaMPI}, and our method. The top and bottom rows depict the synthesized views for two different scenes, with corresponding error maps underneath. The red rectangles highlight specific regions of interest to emphasize differences in synthesis accuracy and visual fidelity.}
    \label{fig:PLFNet_NVS_Results}
\end{figure*}

\subsubsection{Evaluation Metrics}
\label{sect:experiments_nvs_metrics}
To objectively evaluate the quality of the synthesized views, we employed multiple standard and advanced metrics. These include peak signal-to-noise ratio (PSNR) and structural similarity index measure (SSIM) \cite{SSIM}, as well as multi-scale structural similarity index measure (MS-SSIM) \cite{wang2003multiscale}, pixel- based visual information fidelity (VIFP) \cite{VIFP}, deep image structure and texture similarity (DISTS) \cite{dists}, and learned perceptual image patch similarity (LPIPS) \cite{lpip2018}. These metrics capture not only pixel-level errors but also perceptual differences, providing a holistic view of the synthesis quality.

\subsubsection{Results}
\label{sect:experiments_nvs_results}
\Gond{In this section, we compare the view synthesis performance of PVSNet with state-of-the-art methods, namely AdaMPI \cite{han2022single_AdaMPI}, TMPI~\cite{khan2023tiled_tmpi}, and SinMPI~\cite{pu2023sinmpi}. To ensure a fair comparison, all methods were trained on the same dataset, using their respective original implementations and training strategies. \Manu{Additionally, we introduce a lightweight variant called \textbf{PVSNet-Lite}, specifically optimized for maximum inference speed with minimal quality degradation. This version uses a simplified rendering network and reduced feature dimensionality to significantly boost FPS, making it ideal for latency-critical applications. The details of this PVSNet-Lite will be available in our open sourced code\footnotemark[1]. }}

\textbf{Quantitative Image Quality:} The quantitative results for view synthesis across different datasets and resolutions are presented in Table~\ref{tab_nvs}. Our proposed PVSNet consistently outperforms AdaMPI~\cite{han2022single_AdaMPI} on the Blender dataset and achieves competitive results on the COCO~\cite{caesar2018coco} dataset. Our method remains competitive across a range of metrics, especially on PSNR and SSIM.

Interestingly, we observe a slight performance drop in AdaMPI and PVSNet when increasing the resolution from \(256 \times 256\) to \(512 \times 512\). This highlights the challenges of scaling models without increasing their capacity to handle higher-resolution inputs. In contrast to other methods that explicitly rely on depth estimation using the DPT~\cite{MiDaS2022} model, our approach learns scene depth implicitly from RGB input alone, demonstrating the flexibility of our method without the need for pre-computed depth maps.

\textbf{Quantitative Processing Time:} \Manu{To better illustrate the trade off between inference speed and image quality, we present a comparison of all methods in Figure~\ref{fig:Quality_vs_FPS}. The plot shows MS-SSIM versus FPS at both \(256 \times 256\) and \(512 \times 512\) resolutions, clearly highlighting that PVSNet achieves an optimal balance, while PVSNet-Lite pushes the boundary further by significantly improving FPS with only a modest drop in quality.} As shown in Table~\ref{tab_nvs_runtime}, our method achieves an inference time of \Manu{0.0181 seconds (55 fps)} for a resolution of \(512 \times 512\), outperforming all other methods by a factor $>$ \Manu{18}. \Manu{We also report results for PVSNet-Lite in Table~\ref{tab_nvs_runtime}, which achieves even higher frame rates (around 104 FPS at \(512 \times 512\)) with only a marginal drop in visual quality.} This makes PVSNet and \Manu{PVSNet-Lite} highly suitable for real-time applications, where rapid generation of novel views is critical, such as in dynamic video streams or interactive virtual environments. \Manu{A pre-recorded video in our supplementary materials shows the performance in dynamic video streams.}

\textbf{Qualitative Image Quality:} Figure~\ref{fig:PLFNet_NVS_Results} provides a visual comparison of the view synthesis results for different methods on the COCO dataset. Each method’s output is accompanied by error maps, allowing for a more detailed analysis of their accuracy.

\begin{table*}[ht]
  \centering
  \caption{LF Reconstruction: Quality, best values in bold, second best in \textit{italics}. Arrows indicate the better direction. We show mean scores for whole test dataset.}
  
  \label{tab_planar_lf_results}
  \begin{tabular}{l*{10}{c}r}
     \hline
      Dataset                       & Method        & PSNR $\uparrow$   & SSIM $\uparrow$   & MS-SSIM $\uparrow$    & VIFP $\uparrow$   & DISTS $\downarrow$    & LPIPS $\downarrow$\\
    \hline
    Flowers~\cite{Srinivasan2017}   & NoisyLFRecon  & \textit{39.9500}  & \textbf{0.9763}   & \textbf{0.9932}       & \textbf{0.9285}   & \textbf{0.0389}       & \textbf{0.0190} \\
                                    & DGLF          & 35.6194           & 0.8773            & 0.9301                & 0.6021            & 0.1556                & 0.1431 \\
                                    & DALF          & 37.3006           & 0.8941            & 0.9589                & 0.7362            & 0.1033                & 0.0911 \\
                                    & IR-V          & 37.9034           & 0.9122            & 0.9645                & 0.7324            & 0.0994                & 0.0707 \\
                                    & LFSphereNet   & \textbf{41.3719}  & 0.9461            & 0.9868                & 0.9060            & 0.0812                & 0.0512 \\
                                    & PVSNet        & 37.2606           & \textit{0.9601}   & \textit{0.9914}       & \textit{0.9062}   & \textit{0.0608}       & \textit{0.0229} \\
    \hline
    Stanford~\cite{raj2016stanford} & NoisyLFRecon  & 35.4917           & 0.9572            & 0.9771                & 0.8380            & 0.0718                & 0.0395 \\
                                    & DGLF          & 35.6509           & \textbf{0.9673}   & \textit{0.9897}       & 0.8655            & 0.0685                & 0.0321 \\
                                    & DALF          & 38.2996           & 0.9172            & 0.9750                & 0.7829            & 0.0667                & 0.0402 \\
                                    & IR-V          & \textit{39.4909}  & 0.9440            & 0.9852                & 0.8358            & 0.0628                & 0.0390 \\
                                    & LFSphereNet   & \textbf{40.9830}  & 0.9488            & 0.9797                & \textbf{0.8842}   & \textit{0.0556}       & \textit{0.0300} \\
                                    & PVSNet        & 36.8530           & \textit{0.9601}   & \textbf{0.9927}       & \textit{0.8679}   & \textbf{0.0553}       & \textbf{0.0182} \\
    \hline
    JPEG Pleno~\cite{JPEG}          & NoisyLFRecon  & 36.6275           & \textbf{0.9575}   & 0.9687                & 0.8549            & \textit{0.0718}       & 0.0435 \\
                                    & DGLF          & 32.3387           & 0.7207            & 0.8166                & 0.4246            & 0.1556                & 0.1431 \\
                                    & DALF          & 35.0285           & 0.8257            & 0.9291                & 0.6462            & 0.1095                & 0.0937 \\
                                    & IR-V          & \textit{37.2085}  & 0.9122            & 0.9719                & 0.7878            & 0.0894                & 0.0605 \\
                                    & LFSphereNet   & \textbf{39.2624}  & \textit{0.9429}   & \textbf{0.9879}       & \textbf{0.9069}   & 0.0737                & \textit{0.0425} \\
                                    & PVSNet        & 33.8982           & 0.9106            & \textit{0.9838}       & \textit{0.8683}   & \textbf{0.0590}       & \textbf{0.0261} \\
    \hline
\end{tabular}
\end{table*}

SinMPI~\cite{pu2023sinmpi} struggles with rendering artifacts, particularly in regions with complex textures and geometry, as evidenced by the error maps. These issues are likely a result of its depth estimation and scene representation. TMPI~\cite{khan2023tiled_tmpi} also displays notable tiling artifacts due to its approach of partitioning the scene into multiple planes, leading to visible seams and discontinuities.

In contrast, both AdaMPI and our proposed PVSNet produce the most visually accurate results, with minimal artifacts. Our method’s ability to efficiently handle reflections, occlusions, and fine textures stands out, \Manu{in addition the} inference speed makes it the more practical choice for real-time applications, balancing high-quality synthesis with the demands of time-sensitive environments. \Gond{Additional visual results when running the PVSNet on in-the-wild photos are presented in our supplementary materials video.}


\subsection{Light Field Reconstruction and View Synthesis}
\label{sect:experiments_light_field_reconstruction}
\Gond{In this section, we demonstrate alternative approach for view synthesis where Light Field reconstruction is the intermediate step. We focus on capability of}
proposed PVSNet architecture in reconstructing full light fields from a single input image. 
In this case study,
we target very small baseline of the LFs, and we use a smaller version of PVSNet
by modifying the positional embedding step to only use the coordinate normalization during training and inference. The setup of PVSNet which utilizes positional embedding with only the coordinate normalization has been studied in detail in our ablation study in Section~\ref{sect:ablation_study_positional_encoding}, where we show that this setup is sufficient to handle small baselines.
\subsubsection{Datasets}
\label{sect:experiments_light_field_reconstruction_datasets}

For LF reconstruction, we evaluated our method using three publicly available datasets: Lytro Flowers~\cite{Srinivasan2017}, Stanford Light Field Archive~\cite{raj2016stanford}, and JPEG-Pleno~\cite{JPEG}. These datasets are widely used for benchmarking light field reconstruction algorithms. The extracted light field images had dimensions of 352x512 with a 7x7 angular resolution. 

We trained PVSNet on 75\% of the Lytro Flowers and Stanford datasets, with the remaining 25\% reserved for testing. The JPEG-Pleno dataset, however, was exclusively used for testing to evaluate the generalization capability of our model, as it was not included in the training process.

\subsubsection{Evaluation Metrics}
\label{sect:experiments_light_field_reconstruction_metrics}
We utilize similar evaluation metrics as described in Section~\ref{sect:experiments_nvs_metrics}. Since the light field reconstruction works in very narrow baseline between each sub-aperture image, and has discrete camera locations, the search space is much smaller. Therefore we expect much better scores in these metrics.

\begin{table*}[ht]
\centering
\caption{Comparison of the inference times required for reconstructing a light field of size 352×512x7x7 (LF Reconstruction column) \Gond{and synthesizing novel views of size 352×512 (View Synthesis column). All timings are measured on an RTX 2070 Super. For most methods, view synthesis includes the model’s LF reconstruction time, transferring data to shaders, and blending. By contrast, PVSNet directly predicts the queried views without requiring a separate LF reconstruction or rendering step.}}
\begin{tabular}{|l|c|c|c|}
\hline \hline
\textbf{Method}     & \textbf{Purpose} & \textbf{LF Reconstruction (sec)}& \textbf{View Synthesis (sec)}\\ 
\hline
NoisyLFRecon        & LF reconstruction &   1.0923  &   2.3933\\
DGLF                & LF reconstruction &  1.3062   &   2.6072\\
DALF                & LF reconstruction &  0.0577   &   1.3587\\
IR-V                & LF reconstruction &  0.1342   &   1.4352\\
LFSphereNet         & LF reconstruction &  0.0008   &   1.3018\\ \hline \hline
PVSNet              & View synthesis & 0.0850 &   0.0079\\ 
\hline \hline
\end{tabular}
\label{tab:results_lf_reconstruction_time}
\end{table*}

\begin{figure*}[htbp]
\setlength\tabcolsep{0.5pt}
\centering
\begin{tabular}{cccccc}
 \textbf{NoisyLFRecon} &
 \textbf{DGLF} &
 \textbf{DALF} &
 \textbf{IR-V} &
 \textbf{LFSphereNet} &
 \textbf{PVSNet}\\ 
 \cite{Zhou2021} &
 \cite{Zhou2020} &
 \cite{Cun2019} &
 \cite{Han2022} &
 \cite{gond2023lfspherenet} &
 (Ours) \\
 \includegraphics[width=1.1in]{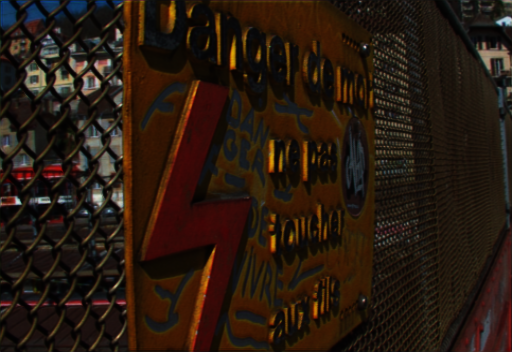} &
 \includegraphics[width=1.1in]{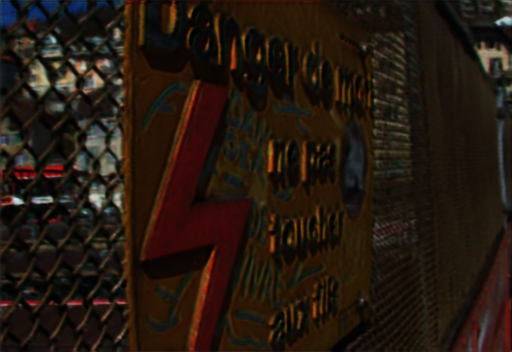} &
 \includegraphics[width=1.1in]{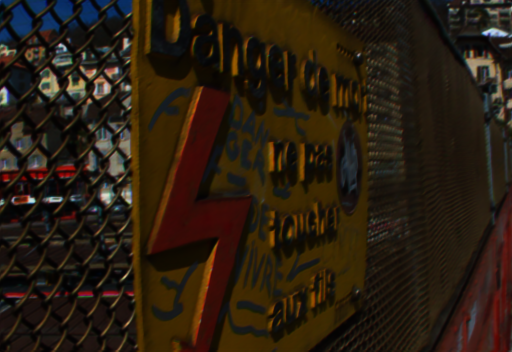} &
 \includegraphics[width=1.1in]{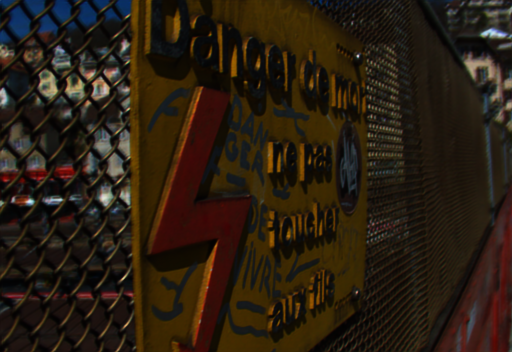} &
 \includegraphics[width=1.1in]{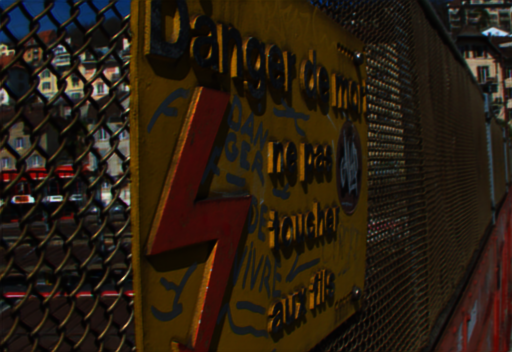} &
 \includegraphics[width=1.1in]{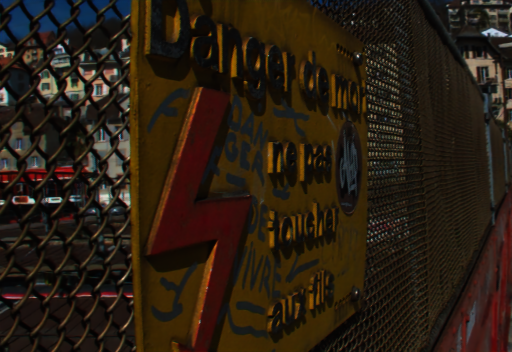} \\
 \includegraphics[width=1.1in]{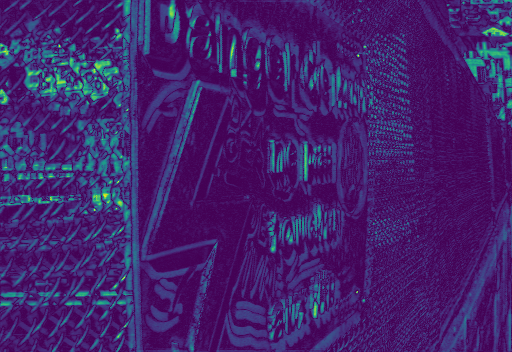} &
 \includegraphics[width=1.1in]{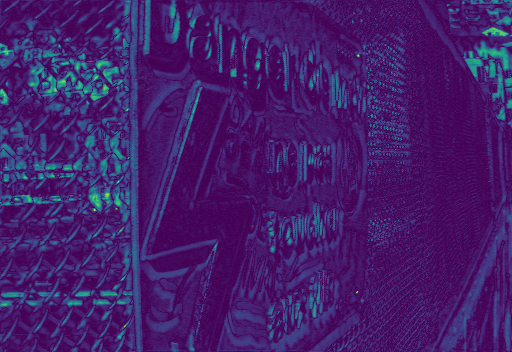} &
 \includegraphics[width=1.1in]{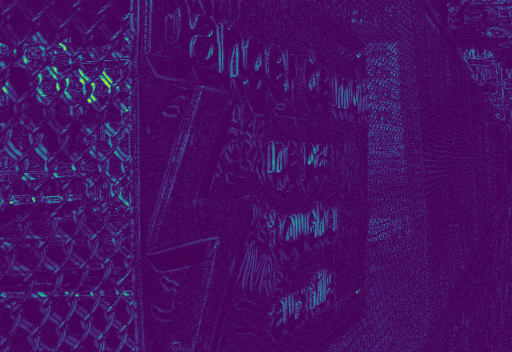} &
 \includegraphics[width=1.1in]{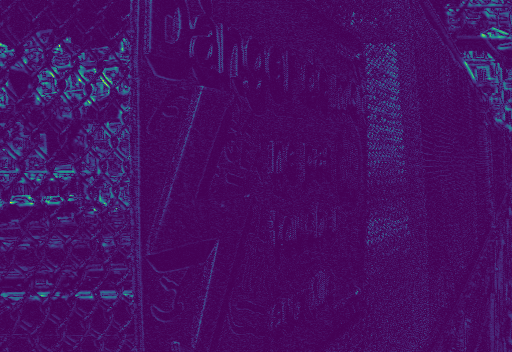} &
 \includegraphics[width=1.1in]{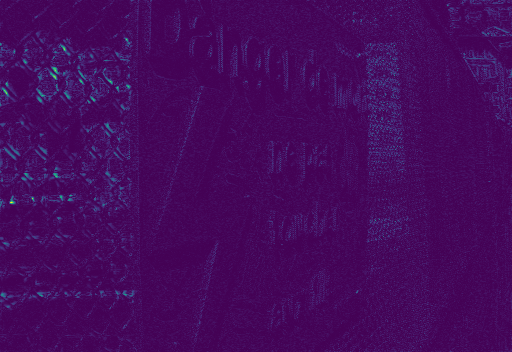} &
 \includegraphics[width=1.1in]{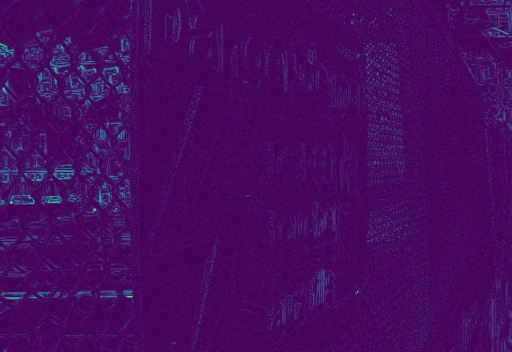} \\
 \includegraphics[width=1.1in]{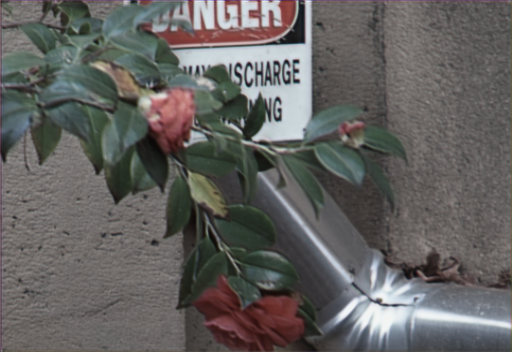} &
 \includegraphics[width=1.1in]{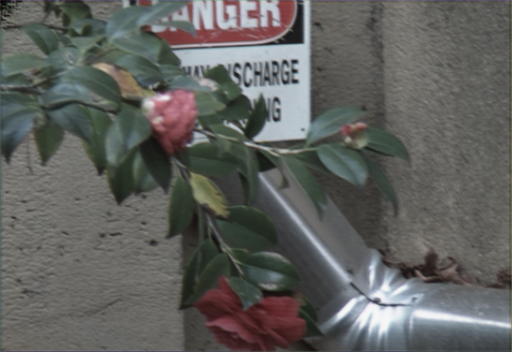} &
 \includegraphics[width=1.1in]{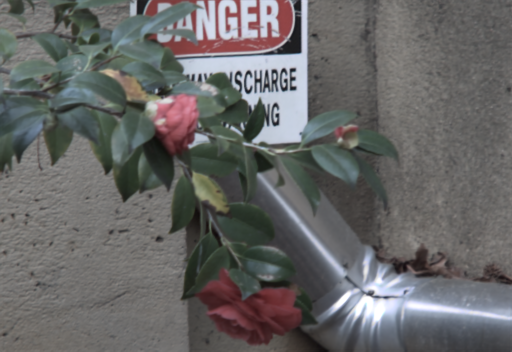} &
 \includegraphics[width=1.1in]{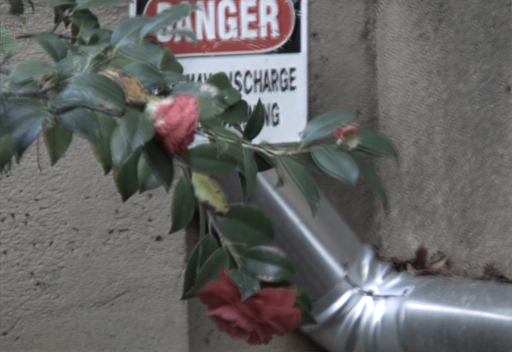} &
 \includegraphics[width=1.1in]{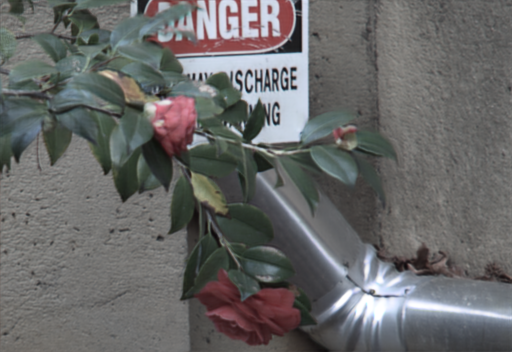} &
 \includegraphics[width=1.1in]{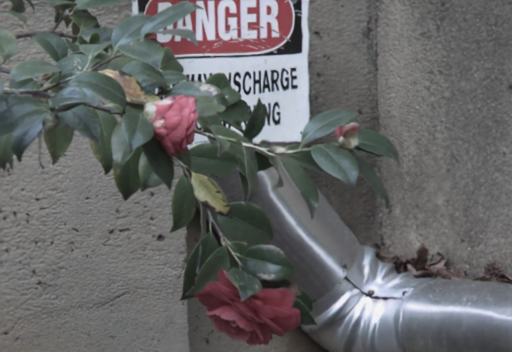} \\
 \includegraphics[width=1.1in]{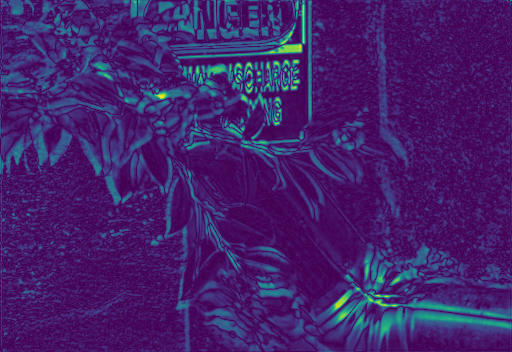} &
 \includegraphics[width=1.1in]{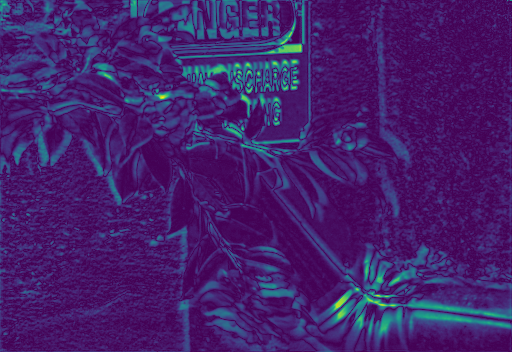} &
 \includegraphics[width=1.1in]{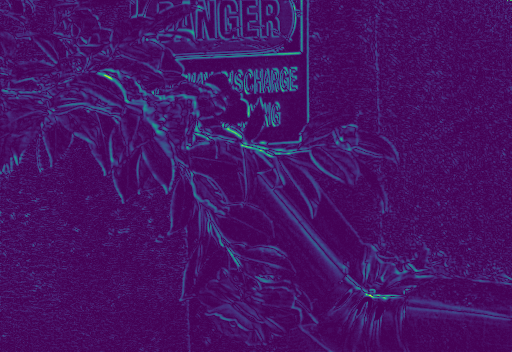} &
 \includegraphics[width=1.1in]{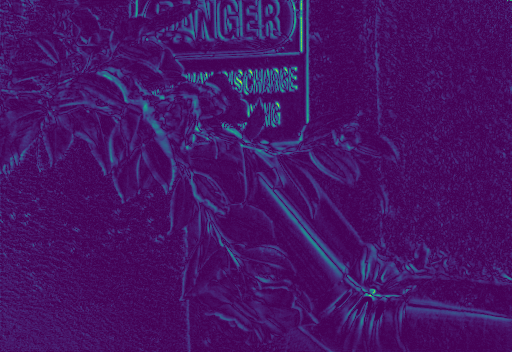} &
 \includegraphics[width=1.1in]{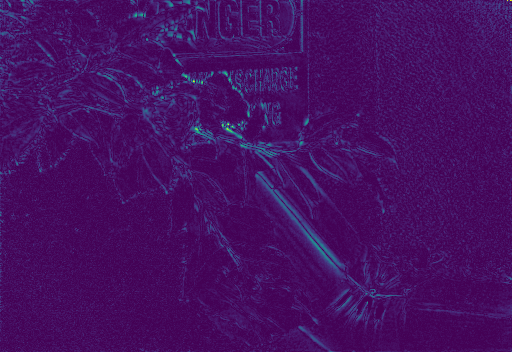} &
 \includegraphics[width=1.1in]{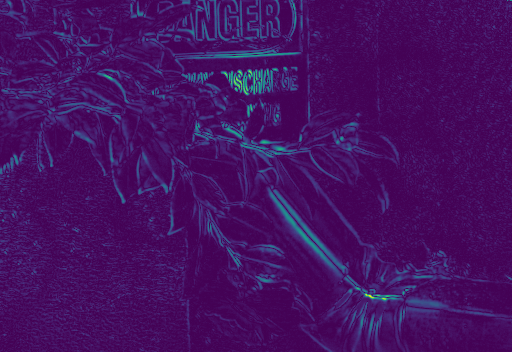} \\
\end{tabular}
\caption{Qualitative results for planar light field reconstruction. 
        We compare results (from left to right) of 
        NoisyLFRecon~\cite{Zhou2021}, DGLF~\cite{Zhou2020}, DALF~\cite{Cun2019}, IR-V~\cite{Han2022}, LFSphereNet~\cite{gond2023lfspherenet} and PVSNet (Ours).
        We have rendered top right sub-aperture image of a 352x512x7x7 planar light field. First two rows shows an image with error map from JPEG Pleno dataset, and last two rows show an image with error map from Stanford dataset.}
\label{fig:qualitative_results_lf_comparision}
\end{figure*}

\subsubsection{Results}
\label{sect:experiments_light_field_reconstruction_results}

\textbf{Quantitative Image Quality:} The quantitative results of our light field reconstruction experiments are presented in Table~\ref{tab_planar_lf_results}. Our proposed PVSNet demonstrates competitive performance compared to state-of-the-art methods, including NoisyLFRecon~\cite{Zhou2021}, DGLF~\cite{Zhou2020}, DALF~\cite{Cun2019}, IR-V~\cite{Han2022}, and LFSphereNet~\cite{gond2023lfspherenet}. All models were trained from scratch using identical datasets for a fair comparison. Publicly available implementations were used for the methods by~\citet{Zhou2021, Zhou2020} and~\citet{Han2022}, while DALF~\cite{Cun2019} was re-implemented in PyTorch according to the original paper.

Although DGLF and LFSphereNet generate the full \(N \times M\) light field in a single pass, both DALF and our PVSNet reconstruct each sub-aperture image iteratively. For the Stanford dataset, which offers fewer training samples compared to Lytro Flowers, LFSphereNet delivered the best results across most metrics, with PVSNet showing competitive performance, particularly excelling in SSIM and MS-SSIM. \Gond{However, we note that while LFSphereNet and the other methods in this comparison were specifically designed for light field reconstruction, PVSNet is optimized for view synthesis within the narrow baseline of light fields. Consequently, PVSNet does not require the blending of sub-aperture views to generate the final synthesized view, setting it apart from the other methods in this comparison.}

\textbf{Quantitative Processing Time:} In terms of computational efficiency, PVSNet demonstrated significant improvements as shown in Table~\ref{tab:results_lf_reconstruction_time}. Our PVSNet achieved an inference time of \textbf{0.0850} seconds for the whole LF reconstruction which included \(7 \times 7 = 49\) forward passes. However, \Gond{PVSNet allows for direct querying of any target view within the light field baseline, enabling significantly faster per-view inference, averaging around 0.0079 seconds (126 fps). This efficiency highlights its suitability for real-time applications. In contrast, methods like LFSphereNet and DALF, despite achieving faster overall light field reconstruction times, necessitate the blending of sub-aperture views using per-view disk based blending algorithm~\cite{overbeck2018system}. This process requires copying the light field data to the GPU buffer and utilizing shaders to blend the views for final synthesis. This additional step introduces an overhead of 1.301 seconds on an RTX 2070 Super, resulting in considerably slower time of 1.3018 seconds for LFSphereNet and 1.3587 seconds for DALF as shown in the last column of Table~\ref{tab:results_lf_reconstruction_time}. Therefore, PVSNet presents a more efficient solution for narrow baseline view synthesis.}

\begin{figure*}[htbp]
\centering
\begin{tabular}{cccccc}
 MLP &
 Normalization &
 Norm+PosEnc &
 Norm+MLP &
 Norm+PosEnc+MLP\\
 \includegraphics[width=1in]{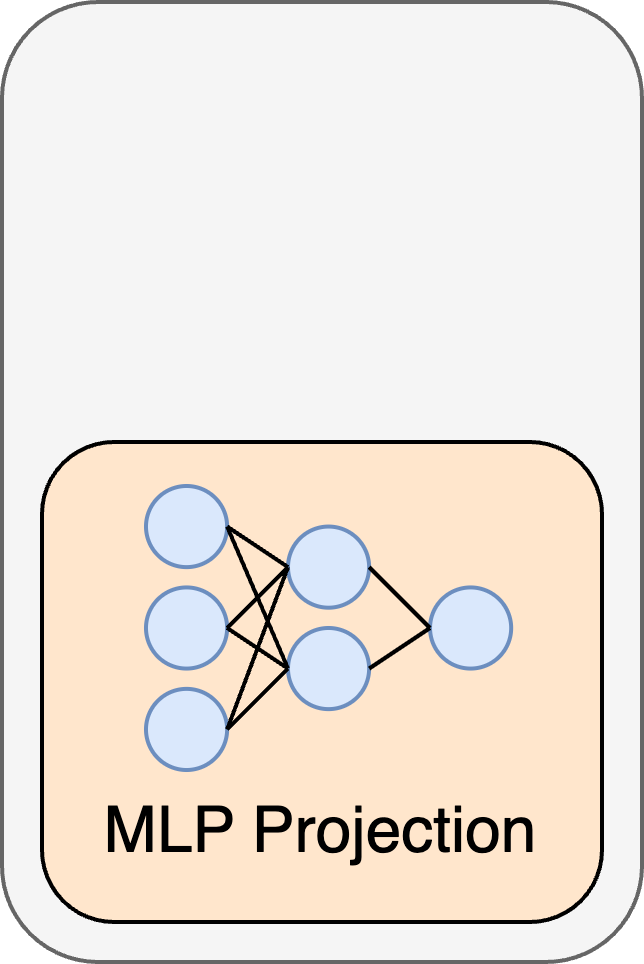} &
 \includegraphics[width=1in]{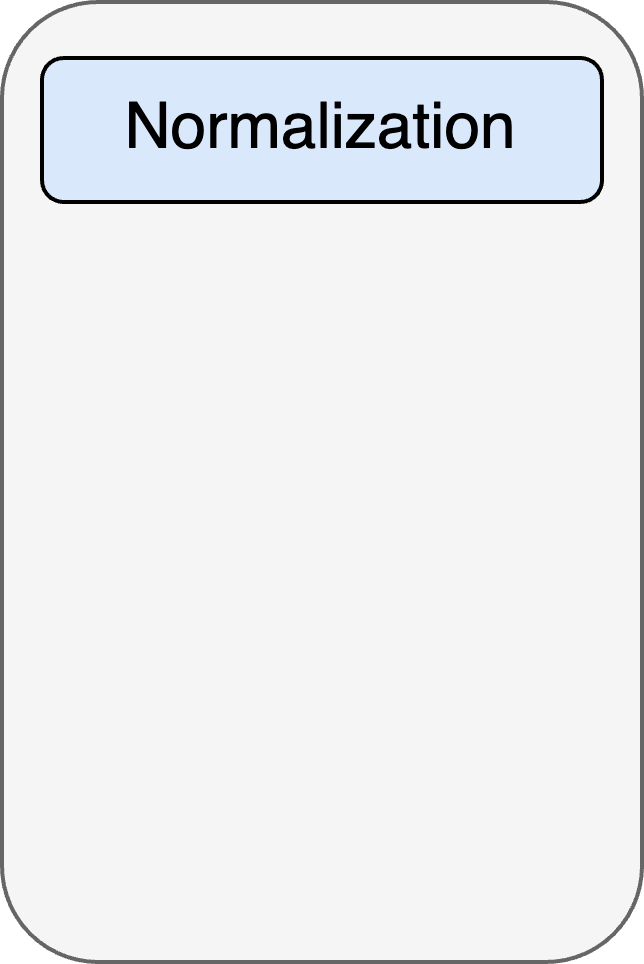} &
 \includegraphics[width=1in]{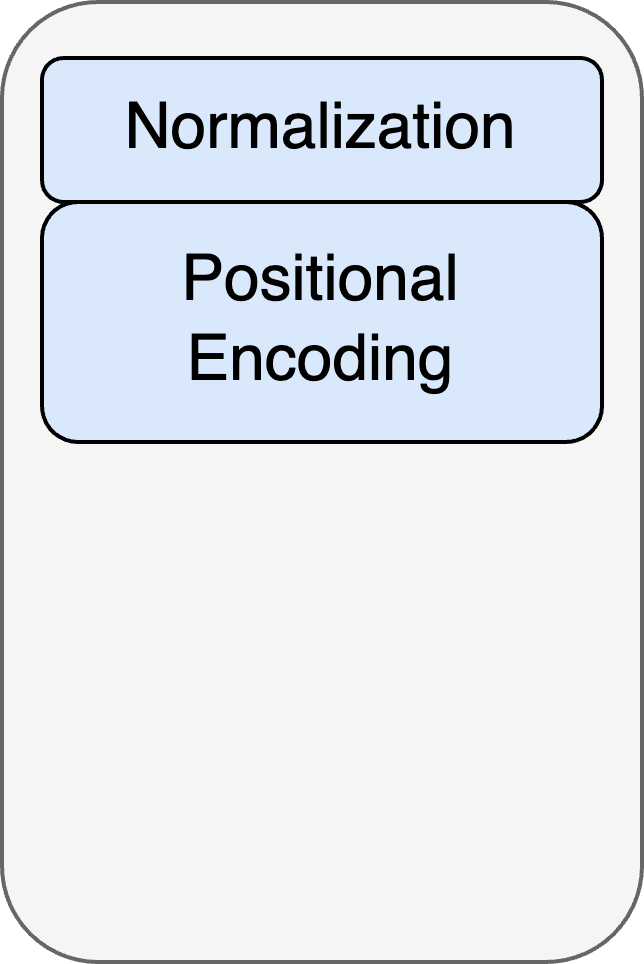} &
 \includegraphics[width=1in]{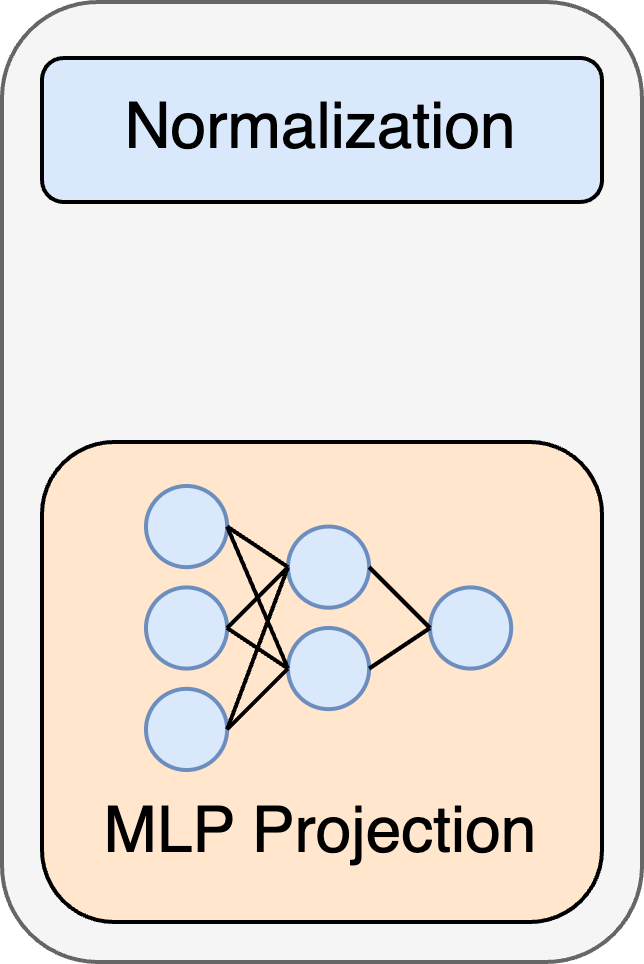} &
 \includegraphics[width=1in]{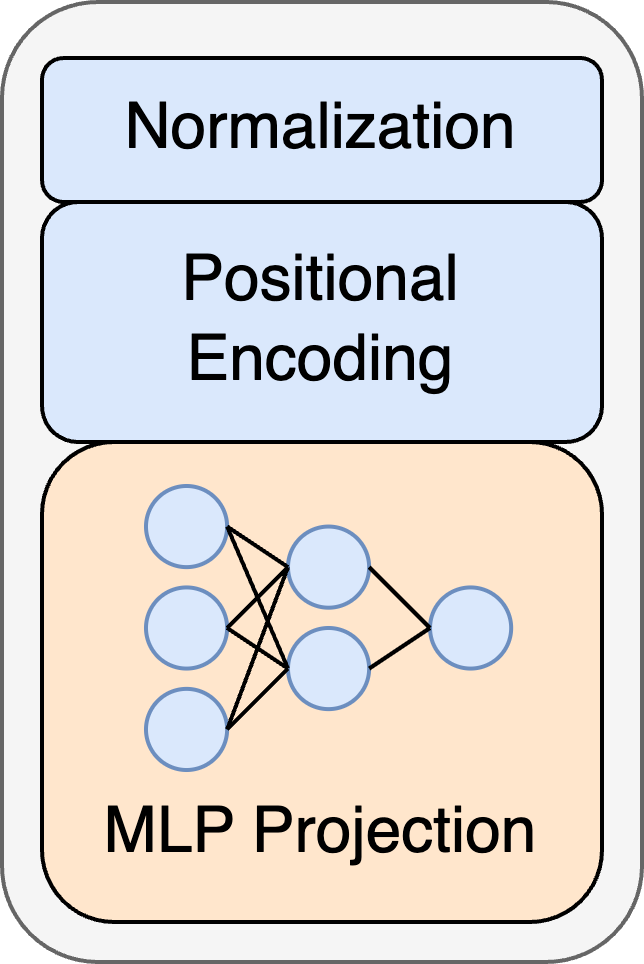}\\
\end{tabular}
\caption{Visual representation of each positional embedding block used in the ablation study. Any selected block can be plugged into the rendering network for training the overall method on selected dataset.}
\label{fig:ablation_study_positional_embedding_block}
\end{figure*}

\begin{table*}[ht]
  \centering
  \caption{View Synthesis Results with Different Positional Embedding Methods: Quality, best values in bold}
  \label{tab_ablation_study_positional_embedding}
  \begin{tabular}{l*{9}{c}r}
     \hline
      Positional Embedding   & PSNR $\uparrow$   & SSIM $\uparrow$   & MS-SSIM $\uparrow$    & VIFP $\uparrow$   & DISTS $\downarrow$    & LPIPS $\downarrow$\\
    \hline
    MLP                      & 23.6780           & 0.8267            & 0.8698                & 0.4577            & 0.1521                & 0.1063 \\
    Normalization            & 25.5882           & 0.8466            & 0.9172                & 0.5673            & 0.1152                & 0.0766 \\
    Norm+Positional Encoding & 24.2862           & 0.8105            & 0.8719                & 0.4675            & 0.1376                & 0.0919 \\
    Norm+MLP                 & 26.7322           & 0.8581            & 0.9366                & 0.6422            & 0.1070                & 0.0888 \\
    Norm+PosEnc+MLP          & \textbf{31.2258}  & \textbf{0.9386}   & \textbf{0.9793}       & \textbf{0.8702}   & \textbf{0.0613}                 & \textbf{0.0275}         \\
    \hline
\end{tabular}
\end{table*}

\textbf{Qualitative Image Quality:} The qualitative evaluation of light field reconstruction is illustrated in Figure~\ref{fig:qualitative_results_lf_comparision}. We compare the visual performance of our proposed PVSNet against state-of-the-art methods for light field views, focusing on the top right corner sub-aperture view. The first two rows depict reconstruction results for the JPEG-Pleno~\cite{JPEG} dataset. In these challenging scenarios, PVSNet demonstrates superior reconstruction quality with fewer visible artifacts, as evident in the error maps. Compared to LFSphereNet~\cite{gond2023lfspherenet}, which also shows strong performance, PVSNet exhibits more accurate details and reduced error magnitudes, highlighting its robustness and generalization capabilities. The subsequent rows display results for the Stanford Light Field Archive~\cite{raj2016stanford} dataset. Here, our method consistently outperforms NoisyLFRecon~\cite{Zhou2021}, which utilizes nine input images for reconstruction. Despite this advantage in quantitative results where the average scores are higher, NoisyLFRecon produces noticeable artifacts and blurring in regions with fine structures. In contrast, PVSNet, which relies on only a single input image, achieves sharper reconstructions with more faithful reproduction of scene geometry and textures. This indicates that PVSNet can effectively capture and synthesize complex light field information even with minimal input data. \Gond{More visual results can be found in
our supplementary materials video, where we show the ability of PVSNet to generate any views within the LF baseline by directly querying the target position.}

Overall, the qualitative results corroborate the quantitative findings, showcasing ability of PVSNet to achieve high-quality reconstructions with low error rates. This performance shows potential for practical applications where high-fidelity light field reconstruction from sparse data is required.



\section{Ablation Study}
\label{sect:ablation_study}
In this section, we present a two-part ablation study to assess the effectiveness of our proposed method, focusing on positional understanding and the impact of different positional embedding techniques. The first part investigates the contribution of different positional embedding strategies to the overall reconstruction quality, highlighting the role of each component in our final model. \Manu{The second part investigates the effect of modifying the structure of the Rendering Network, focusing on the contribution of Encoder~I.} \Manu{Finally, the third} part evaluates the performance of PVSNet under varying sparse input configurations, demonstrating how the number and location of input images influence view synthesis.

\begin{figure*}
    \centering
    \includegraphics[width=0.95\textwidth]{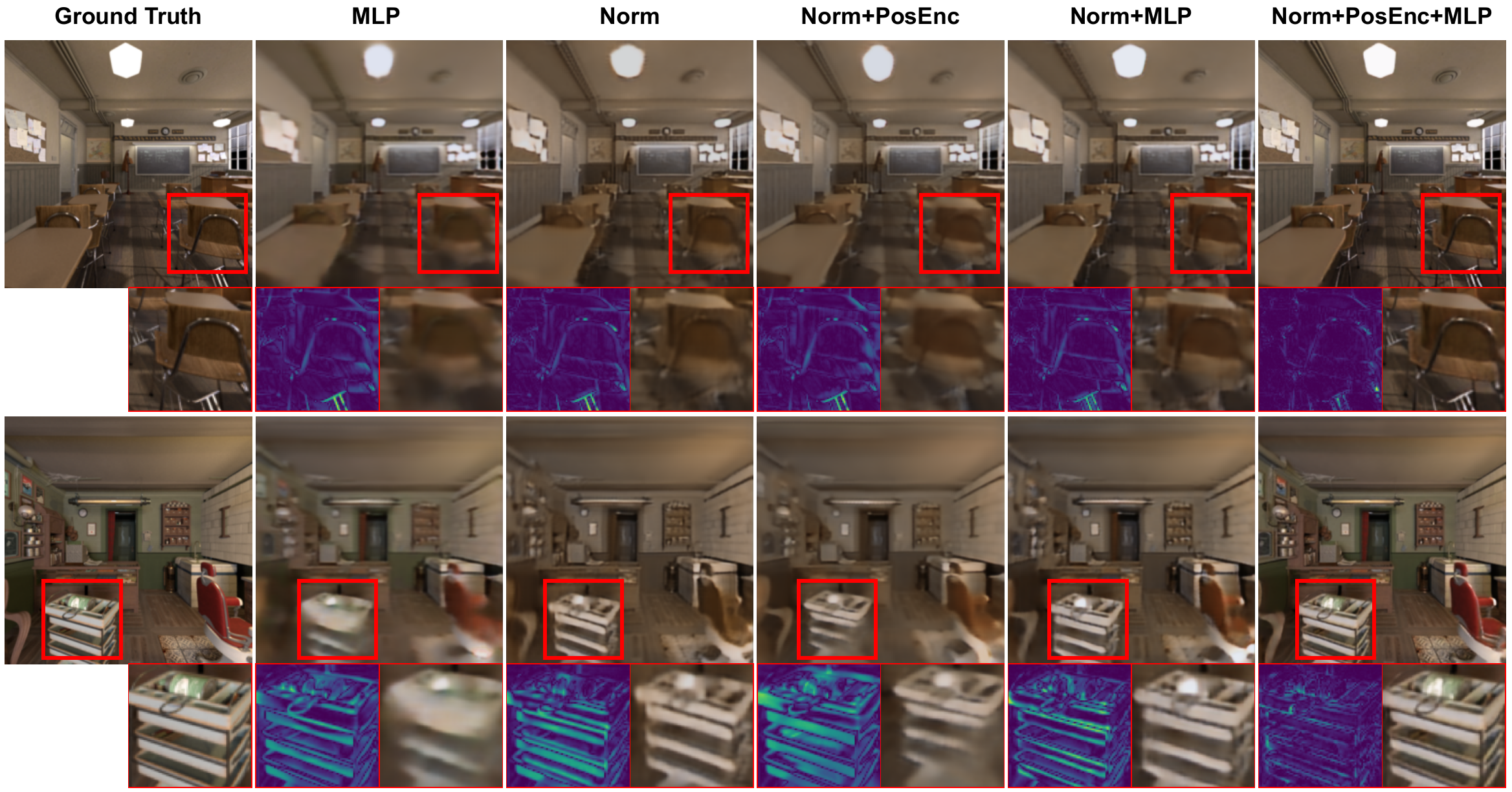}
    \caption{View Synthesis Results with Different Positional Embedding Methods: The red-highlighted areas show zoomed-in regions of the synthesized views alongside their corresponding error maps. The standalone MLP demonstrates the poorest performance, exhibiting significant color inaccuracies and distortions in image features. In contrast, our proposed approach PFLNet, combining normalization, positional encoding, and MLP (Norm+PosEnc+MLP), achieves the most accurate results, preserving both color fidelity and fine image details.}
    \label{fig:PLFNet_Ablation_Study_PosEmbedding}
\end{figure*}

\subsection{Positional Embedding}
\label{sect:ablation_study_positional_encoding}
In the first part of the ablation study, we analyze the impact of different positional embedding strategies on view synthesis. We compare four configurations as shown in Figure~\ref{fig:ablation_study_positional_embedding_block}: (i) MLP only, (ii) normalization only, (iii) normalization with positional encoding, (iv) normalization with MLP, and (v) the full setup with normalization, positional encoding, and MLP. This analysis aims to determine the contribution of each component in enhancing the model’s understanding of positional information. For a fair comparison, we train each variation of the network for same number of epochs. 

The results are summarized in Table~\ref{tab_ablation_study_positional_embedding}. \textit{(i) MLP only:} The model using only the MLP for the positional embedding block without any form of normalization or positional encoding exhibits the weakest performance. The lack of effective positional information results in inability of the rendering network to capture high-frequency variations in the scene, leading to a reconstruction that predominantly represents low-frequency features. This is reflected in a significantly lower average PSNR of \Manu{23.6780 dB,} indicating poor view synthesis quality shown in Figure~\ref{fig:PLFNet_Ablation_Study_PosEmbedding}.
\textit{(ii) Normalization only:} When only coordinate normalization is employed, the network performs reasonably well for small baselines, where the scene geometry and appearance differences between the source and target viewpoints are minimal. However, as the baseline increases, the target viewpoint deviates more significantly from the source, leading to greater variations in scene appearance. Consequently, the model struggles to generalize to these larger baselines, yielding a PSNR of \Manu{25.5882 dB.} This result highlights the limited capacity of normalization alone to effectively handle complex geometric transformations.
\textit{(iii) Normalization with Positional Encoding: } Introducing sinusoidal positional encoding on the normalized coordinates improves ability of the model to account for large geometric variations between source and target views. However, this approach also increases the search space of the rendering network as it must learn to map the given positional embedding vector to a suitable output space. This increased complexity can lead to suboptimal learning, as evidenced by the barbershop scene in Figure~\ref{fig:PLFNet_Ablation_Study_PosEmbedding}, where the rendered image exhibits a noticeable color mismatch. This configuration results in a PSNR of \Manu{24.2862 dB.}

\textit{(iv) Normalization with MLP: } By combining normalization with an MLP, the model benefits from the ability of the MLP to project the normalized coordinates into a higher-dimensional space. This allows the rendering network to better capture both low and high-frequency variations in the scene, making it more effective at handling large baselines. The model achieves a PSNR of \Manu{26.7322 dB,} demonstrating a marked improvement over the previous configurations. This result shows the importance of using an MLP to model complex transformations between source and target viewpoints, especially in scenarios involving significant changes in scene appearance.

\textit{(v) Normalization, Positional Encoding, and MLP: } The final configuration, which incorporates normalization, positional encoding, and an MLP, provides the most comprehensive solution. By applying positional encoding after normalization, the network benefits from a more structured and compact search space, allowing the MLP to more effectively map the positional inputs to a corresponding high dimensional space. This leads to a richer feature representation of the 3D positional information, facilitating the accurate reconstruction of complex novel views. As a result, this configuration achieves the highest PSNR of \Manu{31.2258 dB,} demonstrating its superior performance in synthesizing novel views.

Overall, the results from this ablation study highlight the significant contributions of each component in enhancing the understanding of positional information. The combination of normalization, positional encoding, and MLP provides the most robust and effective solution for accurate and high-quality view synthesis. This combination allows the model to better represent complex geometric and appearance changes, which is essential for synthesizing realistic novel views from a single input image.

\begin{table*}[ht]
  \centering
  \caption{View Synthesis Results with Different Rendering Network Components: Quality, best values in bold}
  \label{tab_ablation_study_rendering_network}
  \begin{tabular}{l*{9}{c}r}
     \hline
      Encoder I & Encoder II   & PSNR $\uparrow$   & SSIM $\uparrow$   & MS-SSIM $\uparrow$    & VIFP $\uparrow$   & DISTS $\downarrow$    & LPIPS $\downarrow$ & FPS $\uparrow$\\
    \hline
    ResNet     & Ours         & 31.2258           & 0.9386             & 0.9793                 & 0.8702             & 0.0613                 & 0.0275 & 62\\
    DINOv2     & Ours         & \textbf{32.5534}  & \textbf{0.9529}    & \textbf{0.9846}        & \textbf{0.8945}    & \textbf{0.0412}        & \textbf{0.0258} & 23\\
     -         & Ours         & 22.9666           & 0.8126             & 0.9288                 & 0.6965             & 0.1251                 & 0.0960 & \textbf{69}\\
    \hline
\end{tabular}
\end{table*}

\begin{table*}[t]
  \centering
  \caption{View Synthesis Results with Sparse Input Mode: Quality, best values in bold}
  \label{tab_ablation_study_sparse_input}
  \begin{tabular}{l*{9}{c}r}
     \hline
      Number of Input Images        & PSNR $\uparrow$   & SSIM $\uparrow$   & MS-SSIM $\uparrow$    & VIFP $\uparrow$   & DISTS $\downarrow$    & LPIPS $\downarrow$\\
    \hline
    2 Images - Random Location       & 25.7561           & 0.8723            & 0.9255                & 0.6328            & 0.1021                & 0.0977 \\
    1 Image - Random Location        & 25.0533           & 0.8298            & 0.9041                & 0.5379            & 0.1227                & 0.1076 \\
    1 Image - Central View          & \textbf{31.2258}  & \textbf{0.9386}   & \textbf{0.9793}       & \textbf{0.8702}   & \textbf{0.0613}                 & \textbf{0.0275}\\
    \hline
\end{tabular}
\end{table*}

\subsection{\Manu{Rendering Network Components}}
\label{sect:ablation_study_rendering_network}
\Manu{In the second part of the ablation study, we analyze the architectural choices made in the rendering network, particularly the use of a dual-encoder setup. The original architecture consists of two encoder branches: \textit{Encoder I:} a ResNet-based feature extractor that processes the input image alone, and \textit{Encoder II:} a custom designed encoder that jointly processes the input image and the positional embedding to generate position aware features. These features are later fused and passed through a decoder to produce the synthesized view.}

\Manu{To understand the contribution of each encoder component, we evaluate three configurations (see Table~\ref{tab_ablation_study_rendering_network}): 
(i) the baseline setup using ResNet for \textit{Encoder I} and our custom block for \textit{Encoder II}, 
(ii) a variant where ResNet is replaced with DINOv2~\cite{DINOv2} to extract more semantically rich features, and 
(iii) a simplified setup where \textit{Encoder I} is entirely removed, relying solely on \textit{Encoder II}.}

\Manu{The results demonstrate that while removing \textit{Encoder I} simplifies the architecture and achieves the highest FPS (69), it significantly degrades synthesis quality across all metrics, including a PSNR drop of over 8 dB. This indicates that the positional features alone are insufficient for capturing fine appearance details without dedicated image encoding. This supports our architectural choice to include an additional image only encoder.}

\Manu{Replacing ResNet with DINOv2 in \textit{Encoder I} further improves quality, achieving the highest PSNR (32.55 dB), SSIM (0.95), and lowest perceptual error (LPIPS of 0.0258). However, this comes at a notable computational cost, reducing inference speed from 62 to 23 FPS. Given our real-time rendering objective, we select ResNet as the backbone for \textit{Encoder I} to maintain a good balance between visual quality and speed.}

\Manu{These findings validate the design choice of using a dual-encoder architecture, where \textit{Encoder I} provides a strong appearance prior, and \textit{Encoder II} injects pose specific spatial cues. The two branches complement each other, leading to sharper, more consistent novel views compared to relying solely on position-aware embeddings.}

\subsection{Sparse Input Mode}
\label{sect:ablation_study_sparse_input}
The third part of the ablation study explores how the performance of PVSNet is affected by varying the number and position of input images for view synthesis. We evaluate three configurations: (i) a single central image as input, (ii) a single image from a random location within the cubical volume, and (iii) two images from random locations. These setups allow us to understand how well the model generalizes to sparse input scenarios, particularly in cases where inputs are not centrally aligned.

Table~\ref{tab_ablation_study_sparse_input} presents the quantitative results of this study. As expected, the single central view input yields the best performance across all metrics, with a PSNR of \Manu{31.2258 dB} and SSIM of \Manu{0.9386,} indicating that the model performs best when provided with a centrally located reference image. The random input configurations, whether with one or two images, exhibit reduced performance, suggesting that while the model can generalize to inputs from random locations, the quality of the synthesized view degrades significantly with less structured input data. Nevertheless, even with these random inputs, PVSNet achieves acceptable performance, demonstrating its robustness in handling diverse input scenarios.

\section{Conclusion and Discussions}
\label{sect:conclusion}
This paper presents PVSNet and its efficient position-aware view synthesis framework, which offers a significant leap in the real-time generation of novel views from a single input image and a target position. Our method demonstrates superior efficiency, achieving real-time performance with a high inference speed, making it particularly suitable for interactive applications like immersive telepresence, 360$^\circ$ video with 6DoF and augmented reality (AR). Through extensive testing, PVSNet outperforms several state-of-the-art methods, such as AdaMPI~\cite{han2022single_AdaMPI}.

\textbf{Contribution to Existing Solutions:} PVSNet enhances the field of view synthesis by introducing an effective solution for real-time inference. Compared to existing methods such as Quark~\cite{Flynn_Quark} or AdaMPI~\cite{han2022single_AdaMPI}, which require multi-view inputs or are computationally expensive, PVSNet excels in generating high-quality novel views from a single image. Its efficiency, particularly with large baselines, marks a significant improvement over previous approaches which require multiple images, or struggle with high computational demands or with generating continuous perspectives. While approaches like NViST~\cite{jang2024nvist} and the use of Multi-Plane Images (MPI) have set the foundation for accurate scene representation, they often face limitations when applied in real-time applications due to their memory and computational overheads. PVSNet addresses these challenges by optimizing both computational efficiency and visual quality, thus supplementing existing solutions and opening up possibilities for their deployment in real-time industrial applications.

\textbf{Implications and Impact:} The outcomes of this research hold substantial implications for both the academic community and industrial applications. For researchers, PVSNet provides a more accessible approach to view synthesis, offering a platform for further innovation in the area of real-time view synthesis. The use of positional embeddings for view synthesis could inspire further exploration of more efficient representations using smaller vision transformer models for real-time scene understanding.

For engineers and industrial applications, the real-time nature of PVSNet makes it a powerful tool for sectors such as remote operations, training simulations, and VR experiences. By providing continuous perspective views with minimal latency, this framework could drastically improve user interactions in these fields. Furthermore, the ability to synthesize novel views from a single input image opens up new possibilities for systems with limited computational resources, where traditional multi-view methods might not be feasible.

However, while the method achieves impressive results, there are still areas for improvement. One limitation is the handling of occlusions and complex geometries in highly dynamic environments when the rendering volume is large for novel view synthesis.

\textbf{Future Directions:} As PVSNet serves as a foundation for real-time position-aware view synthesis, future work can extend this approach by exploring the way to use the latent space generated by the model for other scene understanding tasks, such as depth estimation, segmentation and object detection, to further enhance its applicability in diverse environments. Future research could explore ways to further enhance depth estimation or integrate multi-view information in a hybrid manner, balancing computational efficiency with the ability to handle more complex scene interactions. Additionally, expanding its generalization capabilities to handle a wider range of real-world scenes and augmenting its robustness for large-scale deployment will be crucial steps toward achieving broader industry adoption.




\begin{acks}
The work was supported by the European Joint Doctoral Programme on Plenoptic Imaging (PLENOPTIMA) through the European Union’s Horizon 2020 research and innovation programme under the Marie Skłodowska-Curie Grant Agreement No. 956770, by the Knowledge Foundation, Sweden, with grant number 2019-0251, and by Mid Sweden University internal funding. The computations were enabled by resources provided by the National Academic Infrastructure for Supercomputing in Sweden (NAISS), partially funded by the Swedish Research Council through grant agreement no. 2022-06725. We thank the High Performance Computing Center North (HPC2N) at Umeå University for providing computational resources and valuable support during test and performance runs.
\end{acks}

\bibliographystyle{ACM-Reference-Format}
\bibliography{sample-base}

\end{document}